\RequirePackage{fix-cm}
\documentclass[twocolumn]{svjour3}           %
\smartqed  %
\usepackage{graphicx}
\usepackage{amsmath,amssymb} %
\usepackage{soul}
\usepackage{subfig}
\usepackage{tabularx} %
\usepackage[pagebackref=true,breaklinks=true,colorlinks,bookmarks=false]{hyperref}
\usepackage{booktabs}
\usepackage{pifont}
\usepackage{bold-extra}
\usepackage{dirtytalk}
\usepackage{mathtools}
\usepackage[numbers,sort]{natbib} %

\usepackage[normalem]{ulem}

\usepackage[table]{xcolor}
\usepackage{diagbox} 
\definecolor{badcolor}{rgb}{1.0, 0.94, 0.97}
\definecolor{goodcolor}{rgb}{0.94, 1.0, 0.97}
\definecolor{regularcolor}{rgb}{1., 0.9, 0.9} %
\definecolor{blipcolor}{HTML}{dbe9fc}
\definecolor{clipcapcolor}{HTML}{d5e9d5}
\definecolor{fuchsia}{HTML}{FF00FF}

\def\sepappendix{1}

\newcommand{\gray}[1]{{\textcolor{gray}{#1}}}
\newcommand{\new}[1]{{\textcolor{black}{#1}}}
\newcommand{\newe}[1]{{\textcolor{black}{#1}}}
\newcommand{\white}[1]{{\textcolor{white}{#1}}}

\definecolor{blue}{HTML}{1f77b4}
\definecolor{purple}{HTML}{9467bd}

\def\sepappendix{0} %
\usepackage{multirow}
\usepackage{siunitx}
\begin{document}

\title{Learning text-to-video retrieval from image captioning} %

\author{Lucas Ventura$^{1,2}$ \and Cordelia Schmid$^{2}$ \and G\"ul Varol$^{1}$}
\authorrunning{Ventura, Schmid, Varol}

\institute{
    $^{1}$ LIGM, \'Ecole des Ponts, Univ Gustave Eiffel, CNRS, France \\
    $^{2}$ Inria, ENS, CNRS, PSL Research University, France \\
    \email{\tt\small lucas.ventura@enpc.fr} \\
   {\tt\small \url{https://imagine.enpc.fr/~ventural/multicaps/} }
}

\date{Received: 3 April 2023} 

\maketitle

\begin{abstract} 
We describe a protocol to study text-to-video retrieval
training with unlabeled videos, where we assume 
(i) no access to labels for any videos, i.e.,
no access to the set of ground-truth captions, 
but (ii) access to labeled images in the form of text.
Using image expert models is a realistic scenario given that
annotating images is cheaper therefore scalable, in contrast to expensive video labeling schemes. 
Recently, zero-shot image experts such as CLIP have established
a new strong baseline for video understanding tasks. 
In this paper, we make use of this progress and instantiate
the image experts from two types of models: a text-to-image retrieval model
to provide an initial backbone, and image captioning models
to provide supervision signal into unlabeled videos. 
We show that automatically labeling video frames with image captioning
allows text-to-video retrieval training.
\newe{This process} adapts the features
to the target domain at no manual annotation cost, consequently outperforming the strong zero-shot CLIP baseline.
During training, we sample captions from multiple video frames that \newe{best match} the visual content, and
perform a temporal pooling over frame representations by scoring frames according to their relevance to
each caption.
We conduct 
extensive ablations to provide insights and demonstrate
the effectiveness of this simple framework by outperforming the CLIP 
zero-shot baselines on text-to-video retrieval
on three standard datasets, namely ActivityNet, MSR-VTT, and MSVD.
Code and models will be made publicly available.
\end{abstract}

\section{Introduction}
\label{sec:intro}

\begin{figure}
    \includegraphics[width=.99\linewidth]{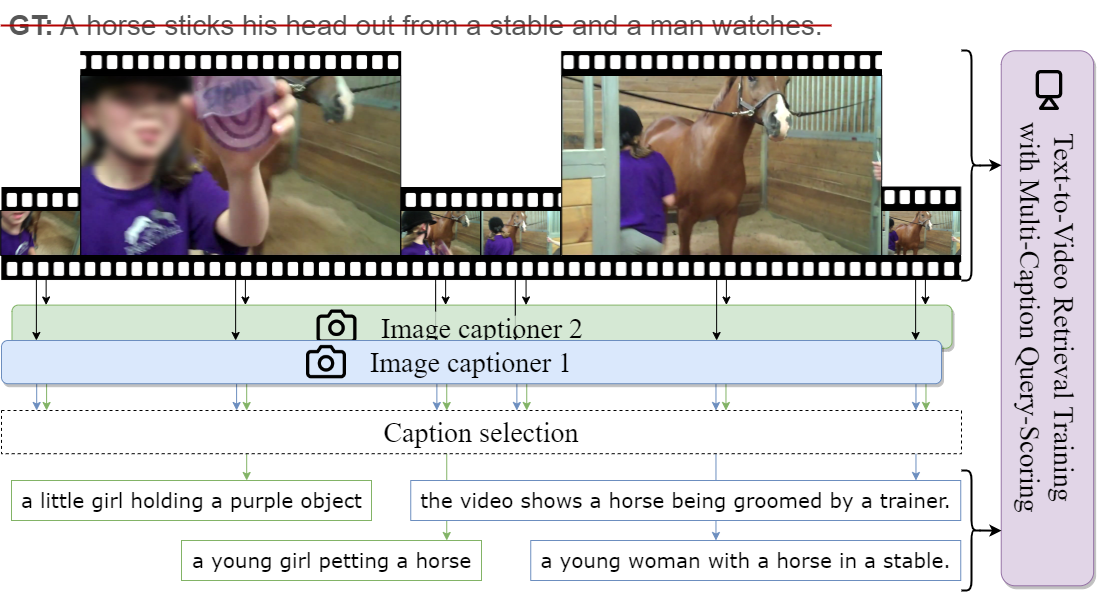}
    \caption{\textbf{Framework:}
        Instead of using the ground-truth video caption, we extract \textit{image}
        captions to automatically label \textit{unlabeled video} frames,
        which we filter to obtain high-quality captions.
        The selected captions from multiple image captioners
        are incorporated into a text-to-video retrieval training
        where each video is paired with multiple caption labels.
    }
    \label{fig:teaser}
\end{figure}

The research on automatic video understanding has witnessed a number
of paradigm shifts recently. Following the rise of neural networks,
the initial question was how to design an architecture to input spatio-temporal
signals~\cite{ng_shortsnippets,tran_c3d}. Given the limited video training data, the focus then shifted
to borrowing parameter initialization from image classification pretraining~\cite{Carreira2017}.
In an attempt to provide \textit{video} pretraining, one
line of work has \newe{made costly efforts to annotate}
video classification datasets~\cite{kinetics400}.
On the other hand, the research community is
moving away from closed-vocabulary recognition training as the progress
in language modeling inspired advances in retrieval of visual data
given open-vocabulary textual input, bridging the gap between symbolic
action categories and describing actions as text~\cite{ChenVideoUnderstanding2021}.
The latest shift was due to the huge scale of labeled image data,
resulting in impressive zero-shot capability of
image-text retrieval models on video action recognition tasks~\cite{clip2021}.
Now, the performance of CLIP-\cite{clip2021}
\new{or BLIP-initialized~\cite{li2022blip}}
image features
(simply averaged over video frames) surpasses most previous works
on a large number of video understanding tasks~\cite{clip4clip2021,ActionCLIP,li2022blip}.
This makes researchers question and rethink where to put their efforts
to improve video modeling. In this study,
we focus on enhancing the \textit{zero-shot} text-to-video retrieval performance of 
CLIP
by making a realistic assumption that we have access to
\textit{image experts}, more specifically an image captioning model.

Fully-supervised methods for video retrieval are limited
due to the high cost of video annotation. Even training with the web-scale
video-text pairs \cite{frozen2021}
do not outperform CLIP image-text pretraining
\cite{Castro2022FitCLIP},
despite the rich descriptions typed manually by humans
with the motivation to sell their videos on stock websites.
On the other hand, methods \newe{that learn}
from unlabeled videos often assume no access to
\textit{any} labels, even for images, with a particular focus on self-supervised
training to use the structure of the data itself as the training signal
\cite{Gordon2020WatchingTW,Yang2020,Feichtenhofer2021}.
In this paper, we ask the question \newe{of} whether an external off-the-shelf
image expert can provide the supervision signal. We explore
the usability of recently released robust image captioners,
namely ClipCap~\cite{ClipCap} and BLIP~\cite{li2022blip}, which benefit from
training with large-scale image-text pairs.~For example, ClipCap uses both CLIP visual pretraining
and GPT-2 language model pretraining~\cite{gpt2019}. When applied
on video frames, we observe that, while noisy, the output texts
contain high-quality descriptions,
which motivates this exploration.

While the idea of using automatic image captions is appealing,
incorporating such \textit{noisy} labels for training introduces
additional challenges. To address this issue,
we first employ a filtering approach where
we select the captions that better describe the frame
by computing the CLIPScore metric \cite{hessel2021clipscore}.
Measuring such cross-modal similarity between the visual
frame and the output text is similar in spirit to the filtering
step in \cite{li2022blip}.
Furthermore, we ensemble multiple image captioners
to obtain a larger pool of labels.
We experimentally validate the benefit of these steps in our ablations.



In this work, we test whether off-the-shelf image
captioning models can serve as an automatic labeling
strategy for video retrieval tasks. We propose a simple framework 
to answer this question. 
Our main baseline, 
as well as our
weight initialization,
is CLIP~\cite{clip2021}.
We finetune
this model
such that
video frame embeddings and the automatic captions
map to the cross-modal joint space after contrastive retrieval
training.
Since one caption may not be representative of the video,
we introduce multi-caption training to effectively use multiple
textual labels per video, by extending the query-scoring method of \cite{max2022Hitchhiker}. \new{This is to overcome
the potential noise in automatic labels, as well as a way to augment data.}
Moreover, since our approach does not
require manual labeling, we can go beyond a single dataset
and combine multiple data sources during training.
This particularly improves performance on smaller datasets.
We demonstrate through experiments that
our approach to pseudo-label unlabeled video frames with image captioning
is a simple, yet effective
strategy that boosts the performance over baselines.


Our contributions are three-fold:
1)~We propose a new simple 
approach to train video retrieval
models using automatic frame captions, which constitute free labels
for supervision (see Figure~\ref{fig:teaser}). To the best of our knowledge, off-the-shelf captioning
has not been used for such objectives \newe{in} prior work \new{at the time of conducting this research\footnote{This paper is an extension of the preliminary work presented in \cite{ventura23multicaps}.}}.
2)~We outperform the zero-shot state-of-the-art CLIP model
on three text-to-video retrieval benchmarks.
3)~We provide extensive ablations about the design choices
on how to select high-quality captions, incorporating
multiple image captioners, temporal pooling with multi-caption query-scoring,
as well as combining multiple datasets.
The code and models will be publicly available.

\section{Related Work}
\label{sec:related}
We briefly overview relevant works on text-to-video retrieval,
self-supervised learning on unlabeled videos, pseudo-labeling, and
captioning.

\noindent\textbf{Text-to-video retrieval.}
Methods for text-to-video retrieval only recently started to train end-to-end neural network 
models~\cite{frozen2021,BridgeFormer}
thanks to (i) the powerful initialization from ViT~\cite{Dosovitskiy2021ViT} and (ii) large-scale video datasets:
noisy HowTo100M data~\cite{miech19howto100m} with ASR-based text supervision from speech,
or more recently the cleaner manually annotated WebVid data~\cite{frozen2021}. 
The progress in text-to-image retrieval \cite{chen2020uniter,clip2021} then triggered advances in text-to-video retrieval. Recent methods employ the CLIP~\cite{clip2021} image backbone
and explore the possibility of adding temporal modeling
(e.g., CLIP2TV~\cite{CLIP2TV}, CLIP4Clip~\cite{clip4clip2021},
CLIP2Video~\cite{CLIP2Video}, CLIP-ViP~\cite{xue2022clipvip}, TS2-Net~\cite{liu2022ts2net}, {ViFi-CLIP~\cite{rasheed2023ViFiCLIP}}).
Their results suggest that the simple averaging of embeddings over frames
remains to be a strong baseline that is difficult to improve on.
Several works have explored fine-grained contrastive learning~\cite{FILIP} for videos
\cite{Jianwei2021TACo,Ma2022X-CLIP},
e.g., considering both frame-word and frame-sentence comparisons \cite{Ma2022X-CLIP}.
Bain~et~al.~\cite{max2022Hitchhiker} \newe{presents} a simple yet effective
method to pool video frame representations with a weighted averaging based on query-scoring.
In this work, we extend this method to use multiple captions instead of a single label per video.
We also use CLIP~\cite{clip2021} as our baseline, as well as our initialization.
Similar to other retrieval methods~\cite{frozen2021,clip4clip2021,Miech2020End-to-end},
we employ a contrastive objective~\cite{InfoNCE}.
Unlike these approaches that assume \new{manually annotated video data
\cite{frozen2021,clip4clip2021,max2022Hitchhiker}
or noisy speech signal \cite{Miech2020End-to-end,xu2021videoclip}},
we obtain our supervision from \textit{automatic} captioning annotations.
\new{In our experiments, we show superior zero-shot performance
over prior models trained on video-text pairs from HowTo100M~\cite{miech19howto100m} or
WebVid~\cite{frozen2021}.}

\noindent\textbf{Self-supervised learning on unlabeled videos.}
A relevant line of work is representation learning on unlabeled videos,
which is often referred to as self-supervised learning. 
In this category,
several works~\cite{Sermanet2017TCN,Sun2019ContrastiveBT,Gordon2020WatchingTW,Yang2020,Feichtenhofer2021} use instance discrimination for videos in a similar fashion with SimCLR~\cite{Chen2020SimCLR} or BYOL~\cite{grill2020byol} in the image setting. The majority of methods also make use of the multimodal nature
of videos, e.g., incorporating the audio signal in the training 
\cite{BraVe2021,morgado2020avid,alwassel2020xdc,piergiovanni2020elo,alayrac2020self}.
A popular approach is to use
the noisy speech signal in uncurated instructional videos such as HowTo100M~\cite{miech19howto100m}. 
The text obtained via ASR is directly considered as the corresponding label,
which is then used within a contrastive objective \cite{miech20endtoend,xu2021videoclip,patrick2021supportset}.
\cite{miech20endtoend} designs a multiple instance training,
VideoCLIP~\cite{xu2021videoclip} performs retrieval-augmented pretraining,
and Support-set~\cite{patrick2021supportset} defines a multi-task captioning objective.
These \new{self-supervised} works may be complementary to our method,
but our focus in this work is different in that
we seek supervision from external image models \new{that provide pseudo-labels,
which can be considered as an alternative route to self supervision}.

\noindent\textbf{Pseudo-labeling.}
Our work is also relevant to pseudo-labeling (or self-labeling) approaches.
Unlike the semi-supervised \cite{Lee2013PseudoLabelT,Sohn2020FixMatchSS,singh2021tcl}
or few-shot \cite{Wang2022LanguageMW} setup considered in these works,
our pseudo-labels do not require any annotations for the problem at hand.
In particular, the concurrent work of
\cite{Wang2022LanguageMW} utilizes image experts to aid video-language learning,
however, requiring a small set of labeled videos.
In a similar fashion, VideoCC~\cite{nagrani2022} exploits image-text
datasets to assign automatic captions to videos for audiovisual retrieval,
but is limited by the finite image captioning dataset source. 
\new{
Our work differs from \cite{nagrani2022} 
by \textit{generating} captions for multiple video frames, rather than retrieving from such a finite set. While these two approaches may potentially be complementary, in our Appendix, we show that nearest neighbor retrieved captions perform worse than generated captions.}

In text-image pretraining, BLIP~\cite{li2022blip} {and BLIP-2~\cite{li2023blip2} employ a} bootstrapping approach
for image captioning, which falls \newe{into} the semi-supervised category, 
\new{i.e.,
they start training with a set of labeled images (whereas we never train on labeled videos). In fact, we employ BLIP as one of our image captioners to obtain automatic
video labels. In our experiments, we also investigate the impact of using BLIP
initialization as opposed to CLIP.}


\noindent\textbf{Captioning.}
There has been increasing interest in the task of generating text to describe a given visual content
\cite{anderson2018bottomup,Lu2018NeuralBT,DonahueHGRVDS15,Venugopalan2015SequenceTS,Park2019AdversarialIF,Chen2014LearningAR,Seo2022EndtoendGP,cho2022fine,yang2023vid2seq}.
\newe{Although} many works focus on integrating object information as additional guidance (e.g., Oscar~\cite{Oscar2020}, VLP~\cite{zhou2019vlp}), such methods perform well on domains similar to that of the object detection
model (e.g., COCO dataset~\cite{coco2014}).
ClipCap~\cite{ClipCap} shows robust performance across datasets
of various domains without making use of an explicit object detection module. Instead,
\cite{ClipCap} makes use of two powerful pretrained models (CLIP~\cite{clip2021} and GPT-2~\cite{gpt2019})
and learns a mapping model between the image features and the language generation.
More recently, BLIP~\cite{li2022blip}, {BLIP-2~\cite{li2023blip2}} and CoCa~\cite{yu2022coca} extend
the contrastive CLIP training by jointly learning image captioning.
Align and tell \cite{Wang2022Align-and-tell} also incorporates
a video captioning head into their text-video retrieval model
during training.
OFA~\cite{wang2022ofa} further supports a variety of image-language tasks
in a unified framework, where captioning can be performed by prompting the visual question answering
model with `What does the image describe?'.
%
Very recently, CapDec~\cite{nukrai2022CapDec} attaches a text decoder on top
of the frozen CLIP image encoder by exploiting text-only data to train an
autoencoder with the CLIP text encoder.

In our work, we employ ClipCap~\cite{ClipCap} and BLIP~\cite{li2022blip}
as our image captioning experts, from which we obtain \newe{the} supervision signal for unlabeled videos.
While both of them are only image-based models, we find that their performance is satisfactory
on video frames. 
The performance of video captioning models are currently behind
those of image captioning approaches, mainly due to limited training data~\cite{Venugopalan2015SequenceTS,Park2019AdversarialIF}. Future work
can explore them as their performance improves.
Recent works of 
ClipVideoCap~\cite{ClipVideoCap}, Lavander~\cite{Li2022Lavander}, CLIP4Caption~\cite{tang2021CLIP4Caption},
{
HiREST~\cite{Zala2023HiREST}, and
TextKG~\cite{gu2023TextKG}
obtain promising results. However,
our setup in this work considers no access to labeled videos.
}

\section{Training with automatic captions}
\label{sec:method}

\begin{figure*} 
	\centering
	\includegraphics[width=.99\textwidth]{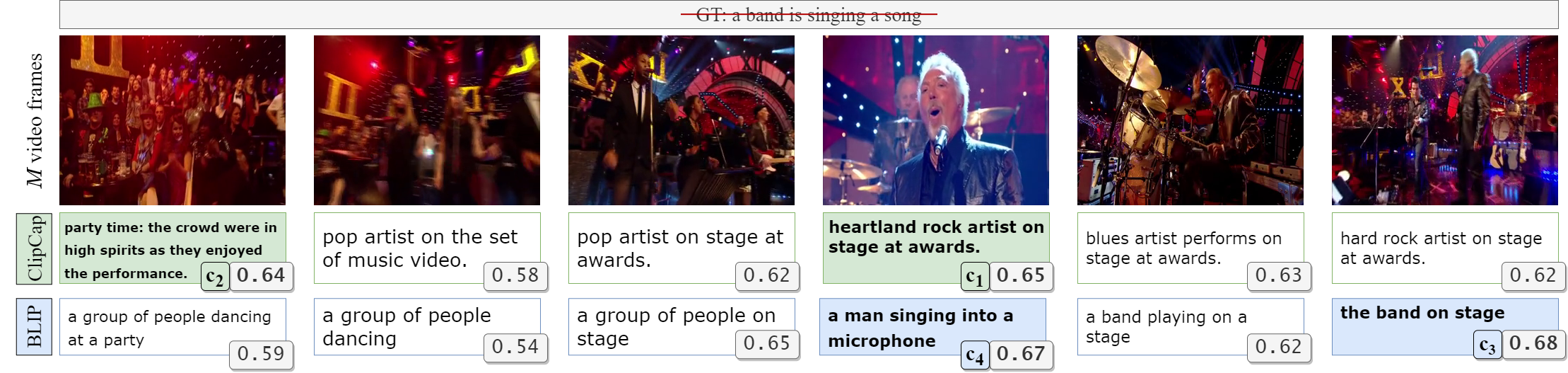}
	{\footnotesize (a)}
	\vspace{-0.2cm}
	\includegraphics[width=.99\textwidth]{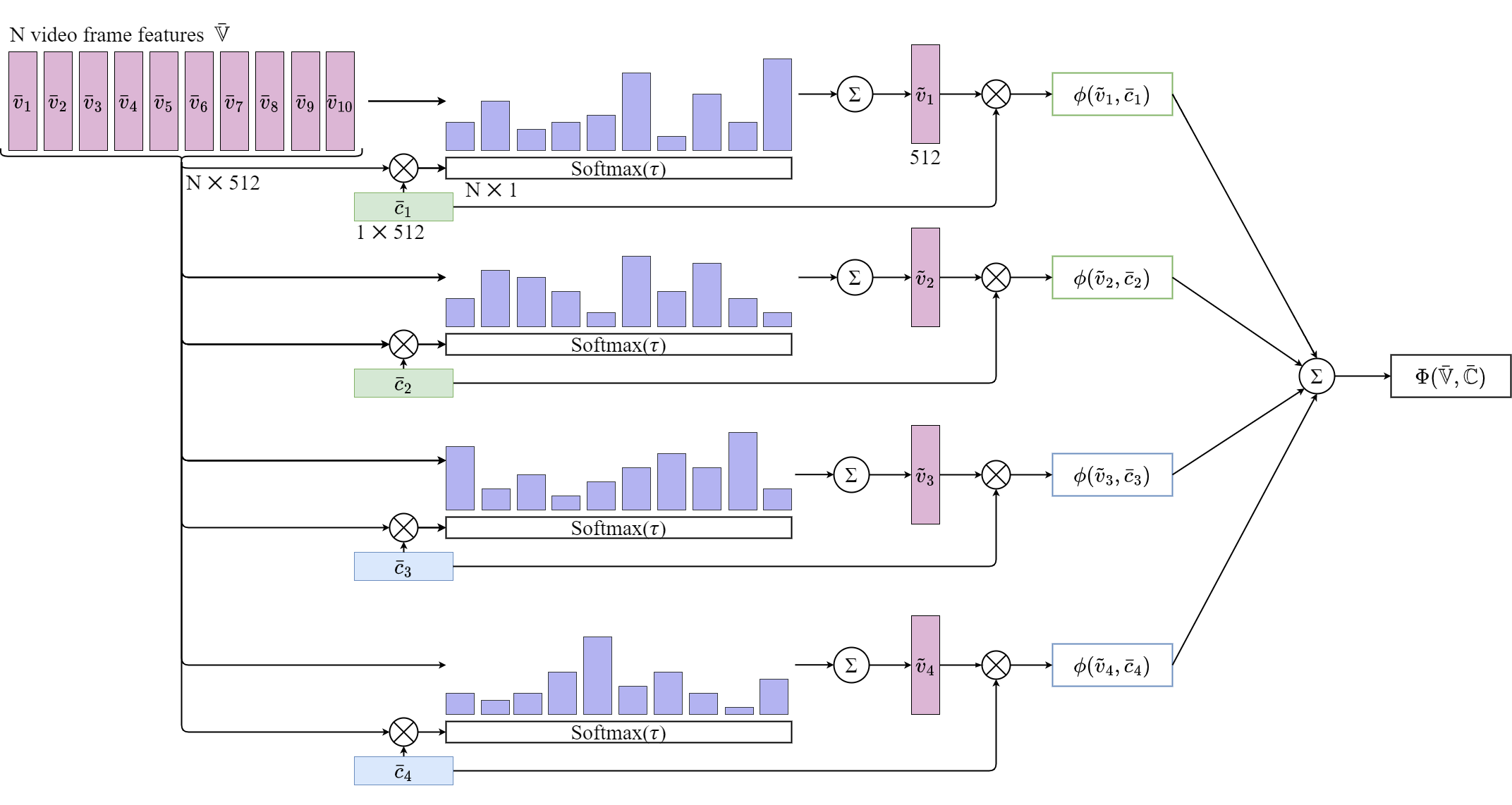}
	{\footnotesize (b)}
	\vspace{-0.2cm}
	\caption{\textbf{Caption selection and multi-caption query-scoring (MCQS):}
		(a) 
		To select the best captions for a given video,
		we first extract image captions from both \colorbox{clipcapcolor}{ClipCap}~\cite{ClipCap} and \colorbox{blipcolor}{BLIP}~\cite{li2022blip} models for $M$ number of frames. We then compute the CLIPScore~\cite{max2022Hitchhiker} (gray box), and finally select Top $K=2$ captions for each captioner: $c_1$ and $c_2$ for ClipCap (highlighted in green), and $c_3$ and $c_4$ for BLIP (highlighted in blue).
		(b) MCQS takes a caption embedding $\bar{c}_l$ and weights the frame embeddings $\bar{v}_1 ... \bar{v}_N$ according to the
		query-scoring temporal poooling function $f_p$ to obtain a
		video representation $\widetilde{v}_l$.
		Finally, 
		we simply average the four similarities obtained with their respective query-scoring.
	}
	\label{fig:method}
\end{figure*}


In this section, we first describe how we
obtain automatic captions for labeling videos,
then present our multi-caption video retrieval training,
and finally\newe{,} give implementation details
for our experimental setup.

The overview of our method is illustrated
in Figure~\ref{fig:method}.
In summary,
we start by constructing a set of
labels for each video, by applying
image captioning models on video frames.
Given these noisy frame-level
captions (from multiple image captioners),
we select the high-quality ones by sorting
them according to their CLIPScore~\cite{hessel2021clipscore}.
We adopt a contrastive video-text retrieval
training using a multi-caption query-scoring
approach, where we incorporate all the selected captions
into the objective. Next, we detail these steps.


\noindent\textbf{Selecting high-quality captions.}
Given an unlabeled training video $v$
consisting of $F$ frames, 
we select $M$ frames from the video ($M \leq F$) and extract captions
using $I$ image captioners to form an initial set of labels
$\mathbb{C}=\{\mathbb{C}_i\}_{i=1}^I$, where $\mathbb{C}_i = \left \{ c_{i1}, c_{i2}, \ldots, c_{iM} \right \}$.
We then obtain $I$ textual descriptions per frame, resulting
in a total of $M \times I$ labels per video.

While we investigate several variants of label formation
from captions in our experiments, our final strategy
is the following.
We select a subset of the initial labels, mainly to eliminate
noisy captions that do not well represent the corresponding video frame.
To this end, we employ CLIPScore~\cite{hessel2021clipscore}
as a way to measure \newe{the} cross-modal
similarity between a caption and its
corresponding frame. For each captioner, we keep the top-$K$
captions ($K < M$) with the highest CLIPScores, which
gives us a remaining $L = K \times I$ labels per video.
We refer to this subset as $\mathbb{C'}$.
Note that some captions are repetitive across frames
due to visual similarity within a video; we therefore
conjecture that such a subset selection does not cause
a significant loss in information.

\noindent\textbf{Contrastive video retrieval objective with multi-caption query-scoring.}
In this work, we employ a relatively standard
vision-language cross-modal training,
where the goal is to find a joint space between
videos and automatic captions.
Given a video $v$, we compute visual embeddings
$\mathbb{\bar{V}} = \{ \bar{v}_n\}_{n=1}^N$
on $N$ video frames ($N \leq F$)
using a visual encoder $f_v : \bar{v}_n \rightarrow \mathbb{R}^d$.
Similarly, we compute textual embeddings with the text encoder $f_t$
from the corresponding set of labels $\mathbb{C'}$
to obtain positive text representations
$\mathbb{\bar{C}} = \{\bar{c}_l\}_{l=1}^{L}$,
where $\bar{c}_l \in \mathbb{R}^{d}$
(with the same embedding dimension as $\bar{v}_n$).
To obtain a single video embedding, we perform temporal pooling over video frame representations.
Inspired by the query-scoring introduced by~\cite{max2022Hitchhiker},
our pooling depends on the text representation, simply through
weighted averaging, where frame weights are proportional to
their similarity with the text. The pooled video embedding 
is then compared against the text to obtain a single similarity.
Differently
from \cite{max2022Hitchhiker}, we have multiple texts $\bar{c}_l$.
We therefore apply query-scoring multiple times, 
and obtain multiple similarities, which we combine by a simple mean
operation (experiments with weighted mean do not yield improvements\newe{;}
see Section~\ref{subsec:ablation}).
More formally,
\begin{equation}
    \Phi(\mathbb{\bar{V}}, \mathbb{\bar{C}}) = \frac{1}{L}\sum_{l \in L}{\phi(\widetilde{v}_l, \bar{c}_l)}, \quad \text{where } \widetilde{v}_l=f_{p}(\mathbb{\bar{V}}, \bar{c}_l),
    \label{eq:mcqs}
\end{equation}
represents a similarity between a set of video frame embeddings $\mathbb{\bar{V}}$ and a set of caption
embeddings $\mathbb{\bar{C}}$, 
where $\phi(.)$ is the cosine similarity
and $f_{p}$ is the query-scoring \cite{max2022Hitchhiker}
temporal pooling function also inputting the text:
\begin{equation}
    f_{p}(\mathbb{\bar{V}}, \bar{c}_l) = \sum_{n \in N}{w_{n} \bar{v}_n} , \text{where}\quad w_n = \frac{e^{\phi(\bar{v}_n, \bar{c}_l) /\tau}}{\sum_{j \in N}e^{\phi(\bar{v}_j, \bar{c}_l)/\tau}}.
\end{equation}
We set the softmax temperature hyperparameter $\tau=0.1$ in our experiments.

From a batch of $B$ visual-texts pair samples,
$\{ (\mathbb{\bar{V}}_1, \mathbb{\bar{C}}_1), (\mathbb{\bar{V}}_2, \mathbb{\bar{C}}_2), ..., (\mathbb{\bar{V}}_B, \mathbb{\bar{C}}_B) \}$,
we train with a symmetric contrastive loss using InfoNCE~\cite{InfoNCE},
i.e., treating all other samples in the batch as negatives:
\begin{equation} 
    \mathcal{L}_{v2c} = -\frac{1}{B} \sum_{b \in B}\textup{log}\frac{\textup{exp}(\Phi(\mathbb{V}_b, \mathbb{C}_b))}{\sum_{j \in B}\textup{exp}(\Phi(\mathbb{V}_b, \mathbb{C}_j))}
\end{equation}
\begin{equation}
    \mathcal{L}_{c2v} = -\frac{1}{B} \sum_{b \in B}\textup{log}\frac{\textup{exp}(\Phi(\mathbb{V}_b, \mathbb{C}_b))}{\sum_{j \in B}\textup{exp}(\Phi(\mathbb{V}_j, \mathbb{C}_b))}
\end{equation}
\begin{equation}
    \mathcal{L} = \mathcal{L}_{c2v} + \mathcal{L}_{v2c},
    \label{eq:loss}
\end{equation}
The final loss
is the sum of video-to-captions ($\mathcal{L}_{v2c}$)
and captions-to-video ($\mathcal{L}_{c2v}$) retrieval loss terms.
Next, we detail the optimization procedure.


\noindent\textbf{Implementation details.}
We instantiate two image captioners ($I=2$) from
ClipCap~\cite{ClipCap} and BLIP~\cite{li2022blip}
models. ClipCap model is pretrained on the
3M images of the Google Conceptual Captions image-text dataset~\cite{sharma-etal-2018-conceptual},
using a MLP mapping between CLIP~\cite{clip2021} image backbone
and GPT-2~\cite{gpt2019} text generation models.
BLIP jointly trains for retrieval and captioning using
129M images (including a subset of LAION~\cite{schuhmann2021laion})
using a bootstrapping approach. We use the publicly
available model, which is further
finetuned on the COCO dataset~\cite{coco2014}.
Given one captioner, we extract $M=10$ captions per video from equally spaced frames.
We empirically set the number of high-quality captions
to top $K=2$ per captioner (i.e., $L = K \times I = 4$).
{On a single GTX1080 GPU, the captioning cost for ClipCap and BLIP is 0.65 fps and 0.93 fps, respectively.}

We minimize the loss function in Eq.~\ref{eq:loss} using Adam~\cite{Kingma2015Adam} optimizer and a learning rate schedule
with a cosine decay~\cite{Loshchilov2017SGDRSG}
as described in \cite{clip4clip2021}. 
For ActivityNet,
we train on 16 Tesla V100 GPUs for 10 epochs, with initial learning rate $10^{-5}$ and mini-batch size $B=64$.
For MSR-VTT and MSVD,
we train on 4 NVIDIA GeForce GTX 1080 for 10 epochs, with initial learning rate $10^{-4}$ and mini-batch size $B=16$.

The weights of our dual encoder model are initialized from CLIP~\cite{clip2021} pretraining
in all experiments unless explicitly stated otherwise,
both for the image ($f_v$) and the text ($f_t$) encoders. 
The image encoder architecture follows 
ViT-B/16~\cite{Dosovitskiy2021ViT} in all experiments. The text encoder architecture
follows GPT-2~\cite{gpt2019}. Both encoders are Transformer-based~\cite{vaswani2017attention},
\new{operating with an embedding dimensionality of $d=512$.}

We resize the frames to 224 $\times$ 224 resolution before inputting \newe{them to} the model.
We use $N=10$ random frame sampling during training based on
segments as in~\cite{wang2019tsn,frozen2021} (note that these
do not necessarily match the $M=10$ captions).
\new{The resulting spatio-temporal raw video input is of 224 $\times$ 224 $\times$ 10 dimensions. 
Each video frame is independently passed through the image encoder to obtain an embedding dimensionality of 512 using the output corresponding to the \texttt{[cls]} token.
The temporal aggregation is obtained via query-scoring as explained above, i.e., weighted averaging over frames where the weights are obtained as frame-text similarity. The resulting video-level representation is therefore of dimensionality 512.}
During training, we use the
multi-caption query-scoring method in Eq.~\ref{eq:mcqs}.
At test time,
we compute the visual embeddings on the center spatial crop over
10 equally spaced frames. During evaluation, as we only have a single query text, multi-caption query scoring is not possible. We thus evaluate using the regular query-scoring method.



\section{Experiments}
\label{sec:experiments}

We start with Section~\ref{subsec:datasets}
by describing the datasets and evaluation metrics
used to report the results of our experiments.
We then present our ablations in Section~\ref{subsec:ablation},
quantifying the effects of
(i)~the captioning model,
(ii)~caption selection,
(iii)~combining captioners,
(iv)~training with multiple captions per video,
and (v)~combining datasets.
Next, we present \newe{a} state-of-the-art comparison in Section~\ref{subsec:sota},
\new{followed by experiments on BLIP initialization
instead of CLIP in Section~\ref{subsec:backbones}}.
Finally, we provide a
qualitative analysis in Section~\ref{subsec:qualitative},
as well as a discussion on limitations in Section~\ref{subsec:limitations}.

\subsection{Datasets and evaluation metrics}
\label{subsec:datasets}

We conduct experiments on three established benchmarks
for text-to-video retrieval, namely
ActivityNet~\cite{krishna2017dense},
MSR-VTT~\cite{Xu2016msrvtt},
and MSVD~\cite{chen-2011-msvd} datasets. 


 
\textbf{ActivityNet Captions~\cite{krishna2017dense}}
contains 20k YouTube videos.
Videos are segmented into 42k clips with \newe{an} average length of 45s.
We use the 10,009 videos from the training set, 
and evaluate on the ``val1'' split (4917 videos).
Note that we extract equally spaced captions per clip, not per video.

\textbf{MSR-VTT~\cite{Xu2016msrvtt}}
is composed of 10k YouTube videos.
The length of the videos varies from 10s to 32s, with an average of 15s.
We train with the Training-9k split
as in \cite{clip4clip2021,frozen2021,Liu2019UseWY,Jianwei2021TACo},
and report results on the 1k split with single video text-pairs as in \cite{wu2018joint,clip4clip2021}.

\textbf{MSVD~\cite{chen-2011-msvd}}
consists of 1970 videos split into 1200 training, 100 validation, and 670 test videos.
The dataset contains both short videos ($\sim$1s) and long videos ($\sim$60s).
Given the small size of the dataset, 
we train using three different seeds and 
average the results on the test split.

As previously explained, even though these datasets contain
ground-truth captions, we do \textit{not} use them during training
(see experiments in Section~\ref{app:sec:fullysup} on fully-supervised setting).
We report the standard evaluation protocols:
text-to-video (T2V) recall at rank 1 and 5 for all experiments.
Recall at rank $k$ (R@$k$)
quantifies the number of times the correct video
is among the top $k$ results.
Higher recall means better performance.

\subsection{Ablation study}
\label{subsec:ablation}
This work constitutes an exploratory study to test whether 
captions can provide \newe{a} training signal for unlabeled videos. 
The answer is yes; however, there are certain design choices we make. 
Here, we provide ablations to measure the sensitivity to these decisions. 
More specifically, we investigate the effects of 
the captioning model
and the quality of the captions provided to the model,
To further improve the results, 
we make use of multiple captions per video
during training, and \newe{combine} datasets to train a single model.

\setlength{\tabcolsep}{6pt}
\begin{table}
\centering
\resizebox{0.99\linewidth}{!}{
\begin{tabular}{l|ll|ll|ll}
    \toprule
    &  \multicolumn{2}{c|}{ActivityNet} & \multicolumn{2}{c|}{MSR-VTT} & \multicolumn{2}{c}{MSVD} \\
    & R@1 & R@5 & R@1 & R@5 & R@1 & R@5 \\
    \midrule
    CLIP baseline~\cite{clip2021} & 23.4 & 49.3 & 32.8 & 55.7 & 39.4 & 64.6 \\
    \midrule
    Ours w/ OFA~\cite{wang2022ofa}   
        & 27.6  & \textbf{55.6}  & 33.6      & 59.2     & \textbf{41.1} & 67.4 \\
    Ours w/ ClipCap~\cite{ClipCap}   
        & 26.7 & 53.5 & 34.7 & 59.8 & 40.6 & 68.9 \\
    Ours w/ BLIP~\cite{li2022blip}   
        & \textbf{27.9} & {54.2} & \textbf{35.8} & \textbf{60.6} & \textbf{41.1} & \textbf{69.1} \\
    \bottomrule
  \end{tabular}
  }
  \caption{\textbf{Captioning models:} Training with automatic captions obtained with
      OFA~\cite{wang2022ofa}, ClipCap~\cite{ClipCap}, and BLIP~\cite{li2022blip}
      all improve over the zero-shot CLIP baseline~\cite{clip2021} on all three
      text-to-video retrieval benchmarks. BLIP captions result in best performances.
  }
  \label{tab:captioners}
\end{table}

\noindent\textbf{(i) Captioning models.}
The first design choice is on the image captioning model to use.
In Table~\ref{tab:captioners}, we present a comparative study
experimenting with three recent captioning models:
OFA~\cite{wang2022ofa}, ClipCap~\cite{ClipCap} and BLIP~\cite{li2022blip}.
More specifically, we use the best available model checkpoints:
OFA-huge trained with 20M publicly available image-text pairs, 
ClipCap trained with Conceptual Captions, and 
BLIP-Large trained with 129M images, finetuned on COCO.
Best results are obtained with BLIP, 
potentially due to the large amount of pretraining
compared to the other two models.
The results also
demonstrate the effectiveness of using captions to improve over the
strong CLIP baseline~\cite{clip2021}, where we
average video frame embeddings using the frozen CLIP.
Note that this is the same as the mean pooling method used in CLIP4Clip~\cite{clip4clip2021}.
In this experiment, we randomly select one caption out of the two best captions during training. We next assess the influence of this selection.


\setlength{\tabcolsep}{6pt}
\begin{table}
\centering
\resizebox{0.99\linewidth}{!}{
\begin{tabular}{ll|ll|ll|ll}
    \toprule
    Captioner & Caption & \multicolumn{2}{c|}{\new{ActivityNet}} & \multicolumn{2}{c|}{MSR-VTT} & \multicolumn{2}{c}{MSVD} \\
     & selection & \new{R@1} & \new{R@5} & R@1 & R@5 & R@1 & R@5 \\
    \midrule
    \multirow{5}{*}{ClipCap} 
    & Rand(10)      & \new{25.1} & \new{51.9}     & 31.8          & 55.2          & 39.8          & 68.5  \\
    & Middle 1      & \new{25.7} & \new{52.4}     & 34.1          & 56.9          & 38.9          & 67.0 \\
    & Top 1         & \new{26.0} & \new{53.3}     & 34.3          & 58.0          & 40.5          & 68.6  \\
    & Rand(Top 2)   & \textbf{\new{26.7}} & \textbf{\new{53.5}}     & \textbf{34.7} & \textbf{59.8} & \textbf{40.6} & \textbf{68.9} \\
    & Rand(Top 3)   & \new{\textbf{26.7}} & \new{\textbf{53.5}}     & 33.1          & 59.0          & 40.5          & 68.4 \\
    \midrule
    \multirow{5}{*}{BLIP} 
    & Rand(10)      & \new{26.3} & \new{52.7}     & 34.6          & 60.5          & 40.5          & 68.7 \\   
    & Middle 1      & \new{25.7} & \new{52.4}     & 33.2          & 57.8          & 40.1          & \textbf{69.9} \\
    & Top 1         & \new{27.6} & \textbf{\new{54.6}}    & 34.9          & 60.3          & \textbf{41.8} & 68.3 \\
    & Rand(Top 2)   & \textbf{\new{27.9}} & \new{54.2}     &\textbf{35.8}  & \textbf{60.6} & 41.1 & {69.1} \\
    & Rand(Top 3)   & \new{27.8} & \new{54.2}    & 35.6          & 59.5          & 40.9          & 68.2 \\
    \bottomrule
  \end{tabular}
  }
  \caption{\textbf{Caption selection:}
  For both captioners, we compare training with a random caption at each epoch,
  training with only the middle frame caption, and training with different number of Top $K$ captions (best CLIPScore~\cite{hessel2021clipscore}). Using CLIPScore filtering improves over using all the 10 captions or only using the middle one on both datasets. Selecting the Top 2 captions results in overall best performance.}
  \label{tab:clipscore}
\end{table}

\noindent\textbf{(ii) Caption selection.}
Automatically generated captions vary in quality.
We select captions with high image-text compatibility to eliminate
potential noise in our training. 
The above image captioning models do not output a confidence score;
therefore, we use CLIPScore~\cite{hessel2021clipscore} between the generated caption and the corresponding input video frame as a caption quality measure.

In Table~\ref{tab:clipscore}, we evaluate whether such filtering
is beneficial.
In this experimental setup, we train with one caption as the video label.
We experiment with five different variants per captioner:
(a) randomly selecting one of the 10 extracted captions at each epoch,
(b) using only the caption corresponding to the middle frame (i.e., 
same label in all epochs),
(c) using only the best caption (i.e., top 1 based on the CLIPscore metric),
(d) randomly selecting one of the 2 best captions at every epoch,
(e) randomly selecting one of the 3 best captions at every epoch.
The results
support the idea that CLIPScore is an effective
filtering method to keep the highest quality captions.
On \new{all three} datasets, and on both captioners (ClipCap and BLIP),
using the best caption(s) slightly improves over using all the captions or the middle one.
\new{Especially for ActivityNet, where the videos are relatively long, it is expected that the caption of the middle frame may not be representative of the video.}
However, there exists a trade-off between 
the number of captions and their quality. 
With more captions per video\newe{,} we avoid overfitting as this may
serve as data augmentation. On the other hand,
the variance among the caption qualities starts to increase.
We empirically find that taking the best two captions
constitutes a good compromise, yielding a promising performance overall. 
\new{However, the difference between top 1, 2, or 3 (last three rows) is not significant.}


\noindent\textbf{(iii) Combining captioners.}
One way to increase the \newe{number} of captions per video
without decreasing the quality of the captions
is to use the best $K$ captions from each captioner
to form the label set.
In Table~\ref{tab:combining-captioners},
we test this hypothesis by taking two captioners
ClipCap and BLIP, to then ensemble their labels.
The results are slightly better than the performance
of individual captioners on most metrics. 
One can potentially further extend to more captioners $I>2$.

Note that we could also select the top \textit{K} from all the captions combined from both captioners.
This would be equivalent to
taking the best 2 captions out of the 20 (10 per captioner).
However, this leads to poorer results, perhaps due to the different CLIPScore distributions
(slight preference for ClipCap potentially because of the CLIP backbone),
and \newe{the} tendency to output repetitive captions across frames for a given captioner.
We provide further analysis in Section~\ref{app:sec:combining-captioners}.

\setlength{\tabcolsep}{6pt}
\begin{table}
    \centering
    \resizebox{0.99\linewidth}{!}{
    \begin{tabular}{l|ll|ll|ll}
        \toprule
        & \multicolumn{2}{c|}{\new{ActivityNet}} & \multicolumn{2}{c|}{MSR-VTT} & \multicolumn{2}{c}{MSVD} \\
        & \new{R@1} & \new{R@5} & R@1 & R@5 & R@1 & R@5 \\
        \midrule
        C & \new{26.7} & \new{53.5} & 34.7 & 59.8 & 40.6 & 68.9 \\
        B & \textbf{\new{27.9}} & \new{54.2} & 35.8 & 60.6 & 41.1 & 69.1 \\
        \midrule
        C+B &  \new{27.3} & \textbf{\new{54.5}} & \textbf{36.5} & \textbf{61.5} & \textbf{41.7} & \textbf{70.0} \\
        \bottomrule
    \end{tabular}
    }
    \caption{\textbf{Combining two captioners:} 
    We observe slight improvements when using captions from both ClipCap (C) and BLIP (B) over using them individually.
    }
    \label{tab:combining-captioners}
\end{table}

\noindent\textbf{(iv) Multi-caption query-scoring (MCQS).}
So far, we have only used one caption as \newe{a} video label during each
training iteration
(even if this is randomly selected from a pool of 4).
Here, we explore how to effectively combine
multiple captions to get a richer video label,
potentially capturing more global content beyond a single\newe{-}frame caption.
In Table~\ref{tab:query-scoring},
we compare 
multi-caption query-scoring (MCQS)
with single-caption query-scoring (QS)
for the 4 captions from ClipCap and BLIP as before.

We first evaluate the effect of QS for the uniform
mean baselines (i.e., only at test time for the CLIP baseline,
and also at training for one random caption baseline).
Our first observation from Table~\ref{tab:query-scoring} is that QS at evaluation
marginally improves the baselines (33.9 vs 32.8 for CLIP,
37.6 vs 36.5 for Rand on MSR-VTT R@1). Training and evaluating
with QS gives a further boost (38.3 vs 37.6). 

{In the last three rows of
Table~\ref{tab:query-scoring}, we then explore three variants of our approach for using
multiple captions:
a) concatenating captions into a single text and just using vanilla QS,
b) weighted, or c) mean similarity pooling in MCQS.
Simple concatenation significantly decreases the performance on MSR-VTT and MSVD,
probably due to the distribution shift caused by the longer sentences during training (4 sentences during training
vs 1 sentence at evaluation). 
On the other hand, ActivityNet results remain similar or even slightly improve
as the standard evaluation protocol also concatenates ground-truth descriptions at test time \cite{clip4clip2021}.
The mean similarity pooling in MCQS obtains an overall improvement
across datasets, over both CLIP and single-caption baselines.
We observe a decrease in performance when dynamically
weighting the similarities based on the ClipScore (with a softmax temperature of 0.1).
We therefore keep the method simple and use the mean of similarities
when jointly training with multiple captions in MCQS.
}





\setlength{\tabcolsep}{6pt}
\begin{table}
\centering
\resizebox{0.99\linewidth}{!}{
\begin{tabular}{l|cc|ll|ll|ll}
    \toprule
    Caption & \multicolumn{2}{c|}{Temporal pooling} & \multicolumn{2}{c|}{ActivityNet} & \multicolumn{2}{c|}{MSR-VTT} & \multicolumn{2}{c}{MSVD} \\
    pooling & \texttt{train} & \texttt{eval} & R@1 & R@5 & R@1 & R@5 & R@1 & R@5 \\
    \midrule
    \multicolumn{2}{l}{\multirow{2}{*}{CLIP baseline~\cite{clip2021}}} & mean & 23.4 & 49.3 & 32.8 & 55.7 & 39.4 & 64.6 \\
    \multicolumn{2}{l}{\multirow{2}{*}{}} & QS & 23.8 & 50.0 & 33.9 & 57.3 & 38.5 & 64.6 \\
    \midrule
    \multirow{3}{*}{Rand(4)} & mean & mean & 27.3 & 54.5 & 36.5 & 61.5 & 41.7 & 70.0 \\
      & mean & QS & 27.8 & 55.0 & 37.6 & 64.3 & 41.9 & 70.0\\
    & QS & QS & 28.4 & 56.6     & 38.3 & \textbf{64.8} & 42.4 & 70.2 \\
    \midrule
    Concat(4) & QS & QS & \textbf{29.8} & \textbf{57.7} & 27.3 & 50.9 & 35.1 & 62.6 \\
    Weighted(4) & MCQS & QS & 29.0 & 57.0 & 38.6 & 63.2 & 41.5 & \textbf{70.5} \\
    Mean(4) & MCQS & QS & 29.7 & 57.1 & \textbf{39.0} & {64.6} & \textbf{42.5} & 70.1 \\
    \bottomrule
  \end{tabular}
  }
  \caption{\textbf{Multi-caption query-scoring:}
  Using all selected captions during training increases performance over only using one caption.
  The CLIP baseline and the model trained with randomly choosing one the 4 caption labels
  are evaluated with query-scoring (QS) for fair comparison. All models use Top-2 from both captioners (i.e., 4 captions in total from C+B).
  }
  \label{tab:query-scoring}
\end{table}

\noindent\textbf{(v) Training with multiple datasets.}
Given that our framework does not require manually annotated videos,
we are not constrained by the fixed size of a dataset's training split,
and we can train with more data.
In Table~\ref{tab:combining-datasets}, we compare how the performance differs 
when: (i) training and evaluating on the same dataset (Self)
versus
(ii) training with more data by combining multiple datasets (Combined).
The resulting combined training set has the following distribution in terms of 
number of video clips coming from each dataset:
$\sim$79\% ActivityNet, $\sim$19\% MSR-VTT, and $\sim$2\% from MSVD.
\new{The percentages represent the relative contribution of each dataset 
to the combined training set, 
derived from the total number of videos available in each dataset, 
with a uniform sampling approach that leads to a higher representation of ActivityNet due to its larger size.}
Such joint training
improves performance moderately for the two relatively bigger datasets
(ActivityNet and MSR-VTT),
and more significantly for the small MSVD dataset.
In the Appendix
\if\sepappendix1{Section~C.1,} \else{Section~\ref{app:subsec:crossdataset},} 
we also report cross-dataset evaluations (e.g., training with ActivityNet and evaluating on MSR-VTT).
\new{This experiment provides additional insights into the generalizability of our approach across different dataset domains.}
An additional advantage is to obtain a single model
instead of multiple dataset-specific models. Future work 
can exploit including larger scale datasets provided
sufficient computing resources.

\setlength{\tabcolsep}{3pt}
\begin{table}
\centering
\resizebox{0.99\linewidth}{!}{
\begin{tabular}{lll|cc|rc|cc}
    \toprule
    & & Vision & \multicolumn{2}{c|}{ActivityNet} & \multicolumn{2}{c|}{MSR-VTT} & \multicolumn{2}{c}{MSVD} \\
    Method & Data & Backbone & R@1 & R@5 & R@1 & R@5 & R@1 & R@5 \\
    \midrule
    \rowcolor{badcolor}
    CLIP w/ QS~\cite{clip2021} & WiT & ViT-B/32 & 20.8  &  45.5     & 30.7  & 54.0  & 33.6  & 62.7  \\
    \rowcolor{badcolor}
    CLIP w/ QS~\cite{clip2021} & WiT & ViT-B/16 & 23.8  & 50.0  & 33.9  & 57.3  & 38.5  & 64.6 \\
    \midrule
    ActBERT~\cite{Zhu2020ActBERT} & H & ResNet-101
            & -     & -     & 8.6   & 23.4   & -  & - \\
    SupportSet~\cite{patrick2021supportset} & H & R(2+1)D-34
            & \textcolor{white}{0}0.1  & \textcolor{white}{0}0.2   & \textcolor{white}{0}8.7   & 23.0   & \textcolor{white}{0}8.9 & 26.0 \\
    MIL-NCE~\cite{Miech2020End-to-end} & H & I3D
            & -     & -     & \textcolor{white}{0}9.9   & 24.0  & -     & - \\
    VideoCLIP~\cite{xu2021videoclip} & H & S3D
            & -     & -     & 10.4  & 22.2   & -    & - \\
    Frozen\cite{frozen2021} & WebVid & ViT-B/16-time
            & -     & -     & 24.7  & 46.9  & - & - \\
    CLIP4Clip~\cite{clip4clip2021} & WiT & ViT-B/32
            & -     & -  & 31.2 & 53.7      & - & - \\
    VideoCC~\cite{nagrani2022} & WiT+VCC & ViT-B/32
            & -     & -     & 33.7  & 57.9   & -  & - \\
    BLIP \cite{li2022blip} (dual) $\dagger$ & B & ViT-B/16 & 26.3 & 52.5  & 35.7  & 59.2  & 35.2 & 63.3 \\
    \gray{BLIP \cite{li2022blip} (cross-modal)}  & \gray{B} & \gray{ViT-B/16} & \gray{35.6}  &  \gray{60.9}   & \gray{43.3}  & \gray{65.6}  & \gray{40.6}  & \gray{67.9} \\
    \midrule
    \rowcolor{goodcolor}
    Ours (Self) & WiT+PL & ViT-B/16 & 29.7 & 57.0 & 39.0 & 64.6 & {42.5} & {70.0} \\
    \rowcolor{goodcolor}
    Ours (Combined) & WiT+PL & ViT-B/16 & \textbf{30.6} & \textbf{57.9}  & \textbf{39.2} & \textbf{65.1} & \textbf{44.6} & \textbf{71.8} \\
    \bottomrule
  \end{tabular}
  }
  \caption{\textbf{Training on the combination of datasets:} We compare training and evaluating on the same dataset (Self), and training with the three combined datasets (Combined = ActivityNet + MSR-VTT + MSVD), and show that combining datasets removes the need of training three separate models
  and slightly improves the overall performance. We perform favorably compared to the state
  of the art on \textit{zero-shot} retrieval (i.e., not using ground-truth video labels
  in downstream datasets).
  {Colored lines are obtained from our implementation. 
  $\dagger$ denotes results we obtained with the code from \cite{li2022blip}. 
  PL is short for pseudo-labels (using automatic captions).
  H:~HowTo100M, VCC:~VideoCC. B:~COCO+VG+CC+SBU+LAION.}}
  \label{tab:combining-datasets}
\end{table}


\subsection{Comparison with the state of the art}
\label{subsec:sota}
In Table~\ref{tab:combining-datasets}, we summarize
other zero-shot methods reporting performances mainly for MSR-VTT,
and our method performs favorably against the state of the art.
\newe{The} rows that are colored are from our implementation,
\new{in comparable settings (e.g., using QS); uncolored rows correspond to other works. Red rows denote our baselines, green rows show our final models.}
Note that CLIP4Clip~\cite{clip4clip2021} zero-shot version
is similar to our CLIP baseline~\cite{clip2021} since they both
use a frozen CLIP to mean-pool over frame embeddings. \new{One difference is our use of query scoring, which was previously ablated in Table~\ref{tab:query-scoring}.
Another difference may be}
due to different hyperparameters such as the number of frames ($N=10$ in ours vs 12 in
\cite{clip4clip2021}).~Note that in contrast to other works, we have access to the training videos 
{(denoted with PL in Table~\ref{tab:combining-datasets})},
albeit without their corresponding ground-truth labels. On the other hand, some of the competitive methods require an external large source 
of videos such as WebVid~\cite{frozen2021} and VideoCC~\cite{nagrani2022}.
\new{Others rely on noisy speech signal from the extensive HowTo100M data \cite{xu2021videoclip,miech20endtoend,patrick2021supportset,Zhu2020ActBERT}, but their performances remain inferior.}

Among prior works, BLIP~\cite{li2022blip} obtains higher performance than our method
on MSR-VTT and ActivityNet.
However, the BLIP model fundamentally differs from dual encoder approaches
in that BLIP also contains a cross-modal encoder that is used for an additional image-text 
matching (ITM in their paper) as a classification task. The matching score from this classification head is then ensembled 
with the cosine similarity obtained by the dual encoder.
Cross-modal encoders are known to perform better than dual encoders; however, they are less efficient~\cite{Miech_2021_CVPR}. We\newe{,} therefore\newe{,} gray out this line in Table~\ref{tab:combining-datasets} to highlight this difference.
On the other hand, we compute the performance of the BLIP dual encoder, 
by considering only the cosine similarity between the unimodal embeddings (similar in spirit to CLIP). The result is much lower, for example for MSR-VTT 35.7 R@1, i.e., lower than both (i) their ensembled result 43.3 and
(ii) our best model using only a dual encoder 39.2.
\new{We next extend our investigation to evaluate the applicability of our method on this more recent cross-modal BLIP encoder as an intialization instead of CLIP.}




\setlength{\tabcolsep}{6pt}
\begin{table}
\centering
\resizebox{0.99\linewidth}{!}{
\begin{tabular}{lcc|cc|cc|cc}
    \toprule
    Backbone & Cross-modal & Method & \multicolumn{2}{c|}{ActivityNet} & \multicolumn{2}{c|}{MSR-VTT} & \multicolumn{2}{c}{MSVD} \\
    (init.)  & Encoder? && R@1 & R@5 & R@1 & R@5 & R@1 & R@5 \\
    \midrule
    \multirow{2}{*}{CLIP~\cite{clip2021}} & \multirow{2}{*}{No} & Baseline & 23.8 & 50.0 & 33.9 & 57.3 & 38.5 & 64.6 \\
    && Ours & 29.7 & 57.1 & {39.0} & {64.6} & \textbf{{42.5}} & \textbf{70.1} \\
    \midrule
    & \multirow{2}{*}{No} & Baseline & 21.0 & 45.4 & 33.0 & 54.8 & 31.3 & 59.7 \\
    BLIP~\cite{li2022blip} & & Ours & 23.4 & 48.0 & 33.8 & 60.5 & 33.7 & 62.2  \\
    \cmidrule(lr){2-9} 
    w/o COCO& \multirow{2}{*}{Yes} & Baseline & 32.1 & 57.1 & 41.4 & 63.3 & 39.6 & 63.9 \\
    && Ours & 32.5 & 59.3 & 42.1 & 64.0 & 40.2 & 66.3 \\
    \midrule
    \multirow{4.5}{*}{BLIP~\cite{li2022blip}} & \multirow{2}{*}{No} & Baseline & 28.2 & 56.3 & 37.4 & 62.2 & 37.7 & 67.3 \\
    && Ours & 30.7 & 58.3 & 39.4 & 64.4 & 38.2 & 67.6 \\
    \cmidrule(lr){2-9} 
    & \multirow{2}{*}{Yes} & Baseline & \textbf{35.1} & \textbf{60.6} & \textbf{43.5} & 66.3 & 40.6 & 67.9 \\
    &&  Ours & 34.6 & \textbf{60.6} & \textbf{43.5} & \textbf{66.5} & {42.2} & {68.5} \\
    \bottomrule
  \end{tabular}
  }
  \caption{\new{\textbf{Initialization with BLIP:}
  We show the comparison between the baseline versus our finetuning with automatic captions across various settings: CLIP/BLIP initialization, BLIP backbone with/without COCO finetuning, BLIP backbone with only the dual encoder or with subsequently reranking with its cross-modal encoder. 
  Our method demonstrates improvements over the baseline across different initialization settings, but the gain is reduced as the baseline performance increases.
  For fairness, unlike the original BLIP evaluation, we use Query-Scoring (QS) when computing dual encoder similarities.
  }}
  \label{tab:backbones}
\end{table}






\subsection{\new{BLIP initialization}}
\label{subsec:backbones}

\new{
To evaluate the applicability of our method across various model initializations, 
we experiment with additional backbones beyond the primary CLIP model. 
In particular, we incorporate the BLIP model~\cite{li2022blip}, which is available with and without COCO finetuning.
The implementation details of BLIP, are summarized in
\if\sepappendix1{Section~E} \else{Section~\ref{app:sec:blip-implementation-details}} \fi
of the Appendix.
}

\new{
In Table~\ref{tab:backbones},
we compare (a) CLIP and BLIP, (b) two versions of BLIP pretraining, (c) both the efficient dual encoder version and the expensive reranking with the cross-modal version of BLIP as done in \cite{li2022blip}, (d) with/without our finetuning with automatic captions. 
Across all datasets and model configurations, 
we find that our finetuning with automatic captions consistently improves over the baselines,
with the exception of the last two rows.
The improvement is more significant for the CLIP backbone,
than for BLIP where the baseline performance is already close to that of fully-supervised approaches (see
\if\sepappendix1{Table~A.1} \else{Table~\ref{tab:finetuning}} \fi
of the Appendix).
In other words, with greater baseline results of the underlying backbone, the more marginal the performance gains become.}

\new{We further note that the reranking operation with the cross-modal encoder, 
while generally leading to improved performance, is significantly less efficient than using the dual encoder alone. Specifically, in \cite{li2022blip}, an initial retrieval is obtained with the dual encoder, and the top-k ($k=128$) retrieved videos are reranked with the costly cross-modal encoder.
Without the cross-modal encoder, 
the CLIP-based model with our approach demonstrates superior performance (refer to rows with ``No'' under ``Cross-modal Encoder'' in Table \ref{tab:backbones}).
We also clarify that
the BLIP baseline performances for both dual and cross-modal encoder configurations are slightly different when compared to Table~\ref{tab:combining-datasets}, 
due to the incorporation of QS in the evaluation for a fair comparison;
for example, MSR-VTT R@1 shows 37.4 vs 35.7 for the dual encoder and 43.5 vs 43.3 for the cross-modal encoder with and without QS, respectively.
For the cross-modal encoder setup, 
QS is only used at the dual encoder retrieval stage, but not in reranking as the encoder inputs all frames without needing a temporal pooling as in \cite{li2022blip}.
}


We conclude the quantitative experiments by stating that
pseudolabeling text-video retrieval datasets with
image captioning allows finetuning \new{text-to-image backbones} with no manual
annotation cost, which in turn substantially improves, for example 
over the frozen CLIP (e.g., 23.8 vs 30.6 on ActivityNet,
33.9 vs 39.2 on MSR-VTT, and 38.5 vs 44.6 on MSVD in Table~\ref{tab:combining-datasets}).

\begin{figure*}
    \centering
    \includegraphics[width=.99\textwidth]{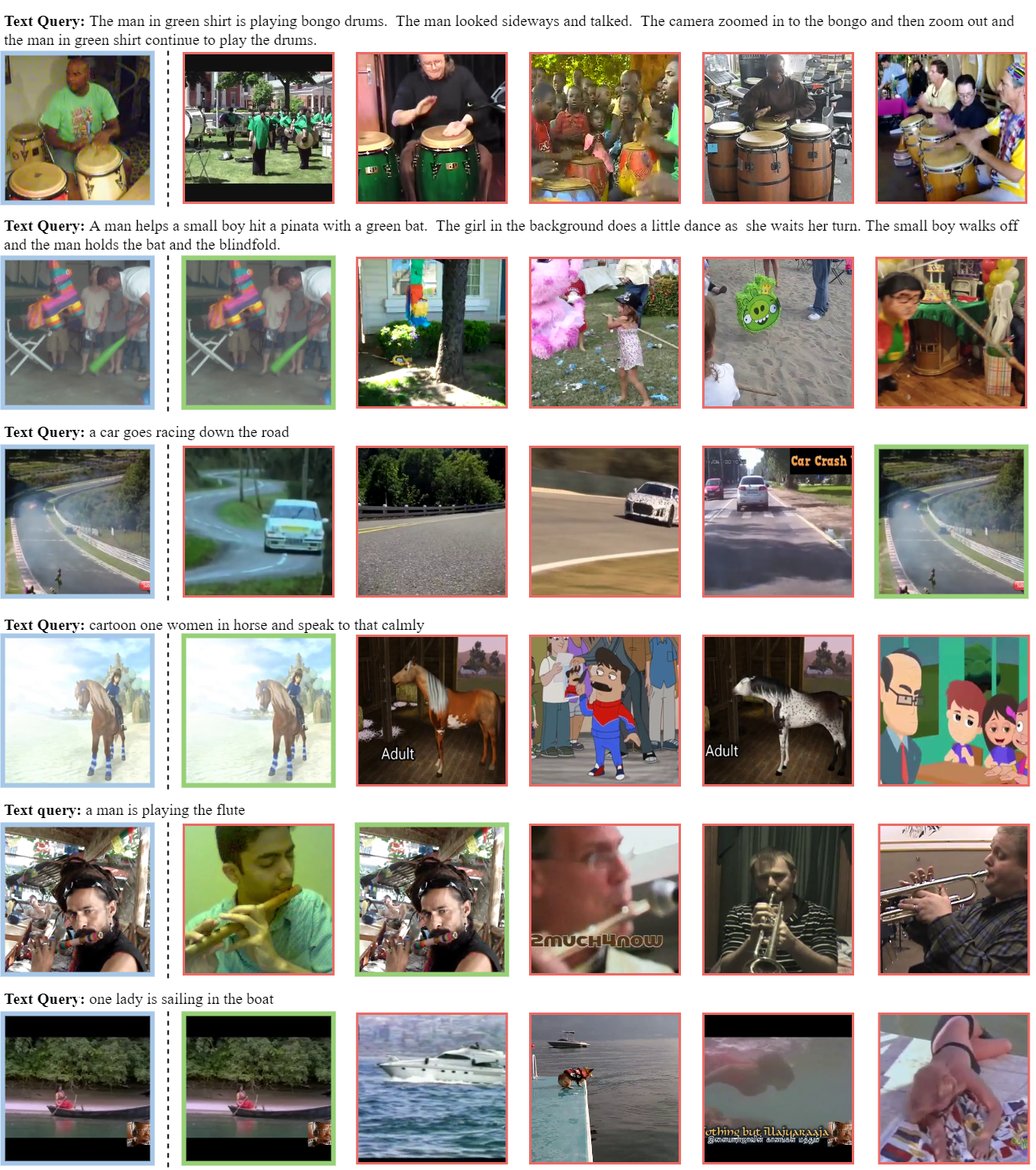}
    \caption{\textbf{Qualitative results:}
    We provide video retrieval results for our best model trained with the combination of the three datasets.
    The examples belong to the test sets of ActivityNet (first two rows), MSR-VTT (third and fourth rows), and MSVD (last two rows). For each example, we show the text query, the ground-truth video (first column, blue border), and
    top 5 retrieved videos from the gallery. Each video is only displayed using the middle frame, with a green border if matches the ground-truth video, or a red border otherwise. Overall, even cases where the correct video
    is not retrieved at the first rank, all the retrieved videos have similar semantic meaning with the text query.
    }
    \label{fig:qualitative-results}
    
\end{figure*}


\subsection{Qualitative analysis}
\label{subsec:qualitative}
In Figure~\ref{fig:qualitative-results}, we illustrate text-to-video results on several examples on all three datasets.
For each test example, we display 
(a)~the textual query,
(b)~the ground-truth video corresponding to the textual query (first column with blue border),
(c)~middle frames of the top 5 retrieved videos (in order from highest to lowest similarity), and
(d)~highlighted green border if the video matches the correct video, or a red border otherwise.
Note that we only visualize the middle frame, which might not be representative
for the overall video. We observe that most of the retrieved videos contain relevant information
to the query text. For example, with the text query: ``\textit{cartoon one women in horse and speak to that calmly}'', all the retrieved videos show cartoons.
Moreover, sometimes even if the correct video is not ranked
in the first position, there may be more than one valid option
(e.g., the text query: ``\textit{a man is playing the flute}'').
We provide
more examples in Section~\ref{app:sec:qualitative}.


\subsection{Limitations}
\label{subsec:limitations}
Here, we discuss several limitations of this work.
First, we note that image captioning does not
necessarily capture the dynamic content of videos.
In particular, some videos may only be recognized
when observing several frames.
Similarly, our temporal pooling approach remains
simple, ignoring the order of frames.
Temporal modeling efforts; however, do not yield
gains for retrieval benchmarks \cite{max2022Hitchhiker}.
As an attempt to incorporate temporal information,
we performed preliminary analysis using
text summarization techniques over the sequence
of captions, but did not obtain consistent improvements (see Section~\ref{app:sec:baselines}).
Another limitation of our experiments is to train on the
videos from the training set of a target
dataset. Even if we do not use their labels,
this setup ensures minimal domain gap.
Future work can leverage large unlabeled video collections
to remove this need.
%
%


\section{Conclusion}
\label{sec:conclusion}
We showed a simple yet effective framework to utilize
an image captioning model as a source of supervision
for text-to-video retrieval datasets.
We demonstrated significant improvements over the strong zero-shot CLIP baseline
with a comprehensive set of experiments.
\new{There are several promising directions for future development. One can}
explore the integration of more image experts beyond captioning,
such as open-vocabulary object detection.
\new{
	The pseudolabeling approach could
be extended to 
a wider variety of video data as mentioned in Section~\ref{subsec:limitations}.
The complementarity of self-supervised representation learning methods
could be investigated to increase the supervision signal in unlabeled videos.
}
Another future direction is to explore 
methods to combine the sequence of image captions into a single video caption.




\bibliographystyle{splncs04}
\bibliography{references}

\begin{thebibliography}{10}
\providecommand{\url}[1]{\texttt{#1}}
\providecommand{\urlprefix}{URL }
\providecommand{\doi}[1]{https://doi.org/#1}

\bibitem{alayrac2020self}
Alayrac, J.B., Recasens, A., Schneider, R., Arandjelovi{\'c}, R., Ramapuram,
  J., De~Fauw, J., Smaira, L., Dieleman, S., Zisserman, A.: {S}elf-{S}upervised
  {M}ulti{M}odal {V}ersatile {N}etworks. In: NeurIPS (2020)

\bibitem{alwassel2020xdc}
Alwassel, H., Mahajan, D., Korbar, B., Torresani, L., Ghanem, B., Tran, D.:
  Self-supervised learning by cross-modal audio-video clustering. In: NeurIPS
  (2020)

\bibitem{anderson2018bottomup}
Anderson, P., He, X., Buehler, C., Teney, D., Johnson, M., Gould, S., Zhang,
  L.: Bottom-up and top-down attention for image captioning and visual question
  answering. In: CVPR (2018)

\bibitem{frozen2021}
Bain, M., Nagrani, A., Varol, G., Zisserman, A.: Frozen in time: A joint video
  and image encoder for end-to-end retrieval. In: ICCV (2021)

\bibitem{max2022Hitchhiker}
Bain, M., Nagrani, A., Varol, G., Zisserman, A.: A {CLIP}-hitchhiker's guide to
  long video retrieval. arXiv  (2022)

\bibitem{banerjee-2005-meteor}
Banerjee, S., Lavie, A.: {METEOR}: An automatic metric for {MT} evaluation with
  improved correlation with human judgments. In: Proceedings of the {ACL}
  Workshop on Intrinsic and Extrinsic Evaluation Measures for Machine
  Translation and/or Summarization (2005)

\bibitem{Carreira2017}
Carreira, J., Zisserman, A.: Quo vadis, action recognition? {A} new model and
  the {Kinetics} dataset. In: CVPR (2017)

\bibitem{Castro2022FitCLIP}
Castro, S., Heilbron, F.C.: {FitCLIP}: Refining large-scale pretrained
  image-text models for zero-shot video understanding tasks. arXiv  (2022)

\bibitem{chen-2011-msvd}
Chen, D., Dolan, W.: Collecting highly parallel data for paraphrase evaluation.
  In: Annual Meeting of the Association for Computational Linguistics: Human
  Language Technologies (2011)

\bibitem{Chen2020SimCLR}
Chen, T., Kornblith, S., Norouzi, M., Hinton, G.E.: A simple framework for
  contrastive learning of visual representations. In: ICML (2020)

\bibitem{Chen2014LearningAR}
Chen, X., Zitnick, C.L.: Learning a recurrent visual representation for image
  caption generation. arXiv:1411.5654  (2014)

\bibitem{chen2020uniter}
Chen, Y.C., Li, L., Yu, L., Kholy, A.E., Ahmed, F., Gan, Z., Cheng, Y., Liu,
  J.: {UNITER}: {U}niversal image-text representation learning. In: ECCV (2020)

\bibitem{cho2022fine}
Cho, J., Yoon, S., Kale, A., Dernoncourt, F., Bui, T., Bansal, M.: Fine-grained
  image captioning with {CLIP} reward. In: NAACL (2022)

\bibitem{devlin-etal-2019-bert}
Devlin, J., Chang, M.W., Lee, K., Toutanova, K.: {BERT}: Pre-training of deep
  bidirectional transformers for language understanding. In: NAACL (2019)

\bibitem{DonahueHGRVDS15}
Donahue, J., Hendricks, L.A., Guadarrama, S., Rohrbach, M., Venugopalan, S.,
  Darrell, T., Saenko, K.: Long-term recurrent convolutional networks for
  visual recognition and description. In: CVPR (2015)

\bibitem{Dosovitskiy2021ViT}
Dosovitskiy, A., Beyer, L., Kolesnikov, A., Weissenborn, D., Zhai, X.,
  Unterthiner, T., Dehghani, M., Minderer, M., Heigold, G., Gelly, S.,
  Uszkoreit, J., Houlsby, N.: An image is worth 16x16 words: Transformers for
  image recognition at scale. In: ICLR (2021)

\bibitem{CLIP2Video}
Fang, H., Xiong, P., Xu, L., Chen, Y.: {CLIP2Video}: Mastering video-text
  retrieval via image {CLIP}. arXiv:2106.11097  (2021)

\bibitem{Feichtenhofer2021}
Feichtenhofer, C., Fan, H., Xiong, B., Girshick, R.B., He, K.: A large-scale
  study on unsupervised spatiotemporal representation learning. In: CVPR (2021)

\bibitem{gabeur2020mmt}
Gabeur, V., Sun, C., Alahari, K., Schmid, C.: {Multi-modal Transformer for
  Video Retrieval}. In: ECCV (2020)

\bibitem{CLIP2TV}
Gao, Z., Liu, J., Chen, S., Chang, D., Zhang, H., Yuan, J.: {CLIP2TV:} an
  empirical study on transformer-based methods for video-text retrieval.
  arXiv:2111.05610  (2021)

\bibitem{BridgeFormer}
Ge, Y., Ge, Y., Liu, X., Li, D., Shan, Y., Qie, X., Luo, P.: Bridgeformer:
  Bridging video-text retrieval with multiple choice questions. In: CVPR (2022)

\bibitem{Gordon2020WatchingTW}
Gordon, D., Ehsani, K., Fox, D., Farhadi, A.: Watching the world go by:
  Representation learning from unlabeled videos. arXiv:2003.07990  (2020)

\bibitem{grill2020byol}
Grill, J.B., Strub, F., Altch'e, F., Tallec, C., Richemond, P.H., Buchatskaya,
  E., Doersch, C., Pires, B.{\'A}., Guo, Z.D., Azar, M.G., Piot, B.,
  Kavukcuoglu, K., Munos, R., Valko, M.: Bootstrap your own latent: A new
  approach to self-supervised learning. In: NeurIPS (2020)

\bibitem{gu2023TextKG}
Gu, X., Chen, G., Wang, Y., Zhang, L., Luo, T., Wen, L.: Text with knowledge
  graph augmented transformer for video captioning. In: CVPR (2023)

\bibitem{hessel2021clipscore}
Hessel, J., Holtzman, A., Forbes, M., Bras, R.L., Choi, Y.: {CLIPScore}: A
  reference-free evaluation metric for image captioning. In: EMNLP (2021)

\bibitem{ChenVideoUnderstanding2021}
Ju, C., Han, T., Zheng, K., Zhang, Y., Xie, W.: Prompting visual-language
  models for efficient video understanding. arXiv:2112.04478  (2021)

\bibitem{kinetics400}
Kay, W., Carreira, J., Simonyan, K., Zhang, B., Hillier, C., Vijayanarasimhan,
  S., Viola, F., Green, T., Back, T., Natsev, P., Suleyman, M., Zisserman, A.:
  The {Kinetics} human action video dataset. arXiv:1705.06950  (2017)

\bibitem{Kingma2015Adam}
Kingma, D.P., Ba, J.: Adam: A method for stochastic optimization. In: ICLR
  (2015)

\bibitem{krishna2017dense}
Krishna, R., Hata, K., Ren, F., Fei-Fei, L., Niebles, J.C.: Dense-captioning
  events in videos. In: ICCV (2017)

\bibitem{Lee2013PseudoLabelT}
Lee, D.H.: Pseudo-label: The simple and efficient semi-supervised learning
  method for deep neural networks. In: ICMLW (2013)

\bibitem{li2023blip2}
Li, J., Li, D., Savarese, S., Hoi, S.: {BLIP-2}: Bootstrapping language-image
  pre-training with frozen image encoders and large language models.
  arXiv:2301.12597  (2023)

\bibitem{li2022blip}
Li, J., Li, D., Xiong, C., Hoi, S.C.H.: {BLIP:} bootstrapping language-image
  pre-training for unified vision-language understanding and generation. In:
  ICML (2022)

\bibitem{Li2022Lavander}
Li, L., Gan, Z., Lin, K., Lin, C.C., Liu, Z., Liu, C., Wang, L.: {LAVENDER}:
  Unifying video-language understanding as masked language modeling. arXiv
  (2022)

\bibitem{Oscar2020}
Li, X., Yin, X., Li, C., Zhang, P., Hu, X., Zhang, L., Wang, L., Hu, H., Dong,
  L., Wei, F., Choi, Y., Gao, J.: Oscar: Object-semantics aligned pre-training
  for vision-language tasks. In: ECCV (2020)

\bibitem{coco2014}
Lin, T.Y., Maire, M., Belongie, S., Bourdev, L., Girshick, R., Hays, J.,
  Perona, P., Ramanan, D., Zitnick, C.L., Dollár, P.: Microsoft coco: Common
  objects in context. In: ECCV (2014)

\bibitem{Liu2019UseWY}
Liu, Y., Albanie, S., Nagrani, A., Zisserman, A.: Use what you have: Video
  retrieval using representations from collaborative experts. In: BMVC (2019)

\bibitem{liu2022ts2net}
Liu, Y., Xiong, P., Xu, L., Cao, S., Jin, Q.: {TS2-Net}: Token shift and
  selection transformer for text-video retrieval. In: ECCV (2022)

\bibitem{Loshchilov2017SGDRSG}
Loshchilov, I., Hutter, F.: Sgdr: Stochastic gradient descent with warm
  restarts. In: ICLR (2017)

\bibitem{Lu2018NeuralBT}
Lu, J., Yang, J., Batra, D., Parikh, D.: Neural baby talk. In: CVPR (2018)

\bibitem{clip4clip2021}
Luo, H., Ji, L., Zhong, M., Chen, Y., Lei, W., Duan, N., Li, T.: {CLIP4Clip}:
  An empirical study of {CLIP} for end to end video clip retrieval.
  arXiv:2104.08860  (2021)

\bibitem{Ma2022X-CLIP}
Ma, Y., Xu, G., Sun, X., Yan, M., Zhang, J., Ji, R.: {X-CLIP}: End-to-end
  multi-grained contrastive learning for video-text retrieval. In: ACMMM (2022)

\bibitem{Miech_2021_CVPR}
Miech, A., Alayrac, J.B., Laptev, I., Sivic, J., Zisserman, A.: Thinking fast
  and slow: Efficient text-to-visual retrieval with transformers. In: CVPR
  (2021)

\bibitem{Miech2020End-to-end}
Miech, A., Alayrac, J.B., Smaira, L., Laptev, I., Sivic, J., Zisserman, A.:
  End-to-end learning of visual representations from uncurated instructional
  videos. In: CVPR (2020)

\bibitem{miech20endtoend}
Miech, A., Alayrac, J.B., Smaira, L., Laptev, I., Sivic, J., Zisserman, A.:
  {E}nd-to-{E}nd {L}earning of {V}isual {R}epresentations from {U}ncurated
  {I}nstructional {V}ideos. In: CVPR (2020)

\bibitem{miech19howto100m}
Miech, A., Zhukov, D., Alayrac, J.B., Tapaswi, M., Laptev, I., Sivic, J.:
  How{T}o100{M}: {L}earning a {T}ext-{V}ideo {E}mbedding by {W}atching
  {H}undred {M}illion {N}arrated {V}ideo {C}lips. In: ICCV (2019)

\bibitem{ClipCap}
Mokady, R., Hertz, A., Bermano, A.H.: {ClipCap}: {CLIP} prefix for image
  captioning. arXiv preprint arXiv:2111.09734  (2021)

\bibitem{morgado2020avid}
Morgado, P., Vasconcelos, N., Misra, I.: Audio-visual instance discrimination
  with cross-modal agreement. arXiv:2004.12943  (2020)

\bibitem{nagrani2022}
Nagrani, A., Seo, P.H., Seybold, B.A., Hauth, A., Manen, S., Sun, C., Schmid,
  C.: Learning audio-video modalities from image captions. In: ECCV (2022)

\bibitem{ng_shortsnippets}
Ng, J.Y.H., Hausknecht, M., Vijayanarasimhan, S., Vinyals, O., Monga, R.,
  Toderici, G.: Beyond short snippets: Deep networks for video classification.
  In: CVPR (2015)

\bibitem{nukrai2022CapDec}
Nukrai, D., Mokady, R., Globerson, A.: Text-only training for image captioning
  using noise-injected {CLIP}. arXiv:2211.00575  (2022)

\bibitem{InfoNCE}
van~den Oord, A., Li, Y., Vinyals, O.: Representation learning with contrastive
  predictive coding. arXiv:1807.03748  (2018)

\bibitem{Park2019AdversarialIF}
Park, J.S., Rohrbach, M., Darrell, T., Rohrbach, A.: Adversarial inference for
  multi-sentence video description. In: CVPR (2019)

\bibitem{patrick2021supportset}
Patrick, M., Huang, P., Asano, Y.M., Metze, F., Hauptmann, A.G., Henriques,
  J.F., Vedaldi, A.: Support-set bottlenecks for video-text representation
  learning. In: ICLR (2021)

\bibitem{piergiovanni2020elo}
Piergiovanni, A.J., Angelova, A., Ryoo, M.S.: Evolving losses for unsupervised
  video representation learning. In: CVPR (2020)

\bibitem{clip2021}
Radford, A., Kim, J.W., Hallacy, C., Ramesh, A., Goh, G., Agarwal, S., Sastry,
  G., Askell, A., Mishkin, P., Clark, J., Krueger, G., Sutskever, I.: Learning
  transferable visual models from natural language supervision. In: ICML (2021)

\bibitem{gpt2019}
Radford, A., Wu, J., Child, R., Luan, D., Amodei, D., Sutskever, I.: Language
  models are unsupervised multitask learners. OpenAI blog  (2019)

\bibitem{rasheed2023ViFiCLIP}
Rasheed, H., Khattak, M.U., Maaz, M., Khan, S., Khan, F.S.: Fine-tuned {CLIP}
  models are efficient video learners. In: CVPR (2023)

\bibitem{BraVe2021}
Recasens, A., Luc, P., Alayrac, J.B., Wang, L., Hemsley, R., Strub, F., Tallec,
  C., Malinowski, M., Patraucean, V., Altché, F., Valko, M., Grill, J.B.,
  van~den Oord, A., Zisserman, A.: Broaden your views for self-supervised video
  learning. arXiv:2103.16559  (2021)

\bibitem{reimers-2019-sentence-bert}
Reimers, N., Gurevych, I.: Sentence-bert: Sentence embeddings using siamese
  bert-networks. In: EMNLP (2019)

\bibitem{schuhmann2021laion}
Schuhmann, C., Vencu, R., Beaumont, R., Kaczmarczyk, R., Mullis, C., Katta, A.,
  Coombes, T., Jitsev, J., Komatsuzaki, A.: {LAION-400M:} open dataset of
  {CLIP}-filtered 400 million image-text pairs. In: Data Centric AI NeurIPS
  Workshop (2021)

\bibitem{Seo2022EndtoendGP}
Seo, P.H., Nagrani, A., Arnab, A., Schmid, C.: End-to-end generative
  pretraining for multimodal video captioning. In: CVPR (2022)

\bibitem{Sermanet2017TCN}
Sermanet, P., Lynch, C., Chebotar, Y., Hsu, J., Jang, E., Schaal, S., Levine,
  S.: Time-contrastive networks: Self-supervised learning from video. In: ICRA
  (2018)

\bibitem{sharma-etal-2018-conceptual}
Sharma, P., Ding, N., Goodman, S., Soricut, R.: Conceptual captions: A cleaned,
  hypernymed, image alt-text dataset for automatic image captioning. In: ACL
  (2018)

\bibitem{singh2021tcl}
Singh, A., Chakraborty, O., Varshney, A., Panda, R., Feris, R., Saenko, K.,
  Das, A.: Semi-supervised action recognition with temporal contrastive
  learning. In: CVPR (2021)

\bibitem{Sohn2020FixMatchSS}
Sohn, K., Berthelot, D., Li, C.L., Zhang, Z., Carlini, N., Cubuk, E.D.,
  Kurakin, A., Zhang, H., Raffel, C.: {FixMatch}: Simplifying semi-supervised
  learning with consistency and confidence. In: NeurIPS (2020)

\bibitem{Sun2019ContrastiveBT}
Sun, C., Baradel, F., Murphy, K.P., Schmid, C.: Contrastive bidirectional
  transformer for temporal representation learning. arXiv:1906.05743  (2019)

\bibitem{tang2021CLIP4Caption}
Tang, M., Wang, Z., LIU, Z., Rao, F., Li, D., Li, X.: {CLIP4Caption: CLIP} for
  video caption. In: ACMMM (2021)

\bibitem{tran_c3d}
Tran, D., Bourdev, L., Fergus, R., Torresani, L., Paluri, M.: Learning
  spatiotemporal features with {3D} convolutional networks. In: ICCV (2015)

\bibitem{vaswani2017attention}
Vaswani, A., Shazeer, N., Parmar, N., Uszkoreit, J., Jones, L., Gomez, A.N.,
  Kaiser, L.u., Polosukhin, I.: Attention is all you need. In: NeurIPS (2017)

\bibitem{ventura23multicaps}
Ventura, L., Schmid, C., Varol, G.: Learning text-to-video retrieval from image
  captioning. In: CVPR Workshop on Learning with Limited Labelled Data for
  Image and Video Understanding (L3D-IVU) (2023)

\bibitem{Venugopalan2015SequenceTS}
Venugopalan, S., Rohrbach, M., Donahue, J., Mooney, R.J., Darrell, T., Saenko,
  K.: Sequence to sequence -- video to text. In: ICCV (2015)

\bibitem{wang2019tsn}
Wang, L., Xiong, Y., Wang, Z., Qiao, Y., Lin, D., Tang, X., Van~Gool, L.:
  Temporal segment networks for action recognition in videos. IEEE Transactions
  on Pattern Analysis and Machine Intelligence  \textbf{41}(11),  2740--2755
  (2019)

\bibitem{ActionCLIP}
Wang, M., Xing, J., Liu, Y.: {ActionCLIP}: {A} new paradigm for video action
  recognition. arXiv:2109.08472  (2021)

\bibitem{wang2022ofa}
Wang, P., Yang, A., Men, R., Lin, J., Bai, S., Li, Z., Ma, J., Zhou, C., Zhou,
  J., Yang, H.: {OFA}: Unifying architectures, tasks, and modalities through a
  simple sequence-to-sequence learning framework. In: ICML (2022)

\bibitem{Wang2022Align-and-tell}
Wang, X., Zhu, L., Zheng, Z., Xu, M., Yang, Y.: Align and tell: Boosting
  text-video retrieval with local alignment and fine-grained supervision. IEEE
  Transactions on Multimedia  (2022)

\bibitem{Wang2022LanguageMW}
Wang, Z., Li, M., Xu, R., Zhou, L., Lei, J., Lin, X., Wang, S., Yang, Z., Zhu,
  C., Hoiem, D., Chang, S.F., Bansal, M., Ji, H.: Language models with image
  descriptors are strong few-shot video-language learners. arXiv  (2022)

\bibitem{xu2021videoclip}
Xu, H., Ghosh, G., Huang, P.Y., Okhonko, D., Aghajanyan, A., Metze, F.,
  Zettlemoyer, L., Feichtenhofer, C.: {V}ideo{CLIP}: Contrastive pre-training
  for zero-shot video-text understanding. In: EMNLP (2021)

\bibitem{Xu2016msrvtt}
Xu, J., Mei, T., Yao, T., Rui, Y.: {MSR-VTT}: A large video description dataset
  for bridging video and language. In: CVPR (2016)

\bibitem{xue2022clipvip}
Xue, H., Sun, Y., Liu, B., Fu, J., Song, R., Li, H., Luo, J.: {CLIP-ViP}:
  Adapting pre-trained image-text model to video-language representation
  alignment. arXiv  (2022)

\bibitem{yang2023vid2seq}
Yang, A., Nagrani, A., Seo, P.H., Miech, A., Pont-Tuset, J., Laptev, I., Sivic,
  J., Schmid, C.: Vid2seq: Large-scale pretraining of a visual language model
  for dense video captioning. In: CVPR (2023)

\bibitem{ClipVideoCap}
Yang, B., Zou, Y.: {CLIP} meets video captioners: Attribute-aware
  representation learning promotes accurate captioning. arXiv:2111.15162
  (2021)

\bibitem{Yang2020}
Yang, C., Xu, Y., Dai, B., Zhou, B.: Video representation learning with visual
  tempo consistency. arXiv:2006.15489  (2020)

\bibitem{Jianwei2021TACo}
Yang, J., Bisk, Y., Gao, J.: {TACo}: Token-aware cascade contrastive learning
  for video-text alignment. arXiv  (2021)

\bibitem{FILIP}
Yao, L., Huang, R., Hou, L., Lu, G., Niu, M., Xu, H., Liang, X., Li, Z., Jiang,
  X., Xu, C.: {FILIP}: Fine-grained interactive language-image pre-training.
  In: ICLR (2022)

\bibitem{yu2022coca}
Yu, J., Wang, Z., Vasudevan, V., Yeung, L., Seyedhosseini, M., Wu, Y.: {CoCa}:
  Contrastive captioners are image-text foundation models. Transactions on
  Machine Learning Research  (2022)

\bibitem{wu2018joint}
Yu, Y., Kim, J., Kim, G.: A joint sequence fusion model for video question
  answering and retrieval. In: ECCV (2018)

\bibitem{Zala2023HiREST}
Zala, A., Cho, J., Kottur, S., Chen, X., Oğuz, B., Mehdad, Y., Bansal, M.:
  Hierarchical video-moment retrieval and step-captioning. In: CVPR (2023)

\bibitem{Zhang2018Cross-Modal}
Zhang, B., Hu, H., Sha, F.: Cross-modal and hierarchical modeling of video and
  text. In: ECCV (2018)

\bibitem{zhou2019vlp}
Zhou, L., Palangi, H., Zhang, L., Hu, H., Corso, J.J., Gao, J.: Unified
  vision-language pre-training for image captioning and {VQA}. In: AAAI (2020)

\bibitem{Zhu2020ActBERT}
Zhu, L., Yang, Y.: {ActBERT}: Learning global-local video-text representations.
  In: CVPR (2020)

\end{thebibliography}

\bigskip \bigskip
\noindent \textbf{\large APPENDIX}
\bigskip
\renewcommand{\thefigure}{A.\arabic{figure}} 
\setcounter{figure}{0} 
\renewcommand{\thetable}{A.\arabic{table}}
\setcounter{table}{0} 

\renewcommand{\thesection}{\Alph{section}}
\setcounter{section}{0}

This appendix provides
{experiments with the fully-supervised setting
(Section~\ref{app:sec:fullysup}),}
results with alternative methods (Section~\ref{app:sec:baselines}),
additional evaluations (Section~\ref{app:sec:evaluations}),
analyses on selecting captions and combining captioners (Section~\ref{app:sec:combining-captioners}),
\new{implementation details about the BLIP initialization experiment (Section~\ref{app:sec:blip-implementation-details}),}
additional qualitative results
(Section~\ref{app:sec:qualitative}),
and a data availability statement (Section~\ref{app:data-availability}).

\section{Fully-supervised setting}
\label{app:sec:fullysup}
While our focus is on the zero-shot setting, where labeled video data is not available, 
it is worth noting that for small-scale datasets, annotation costs may not be prohibitively high allowing for fully-supervised settings.
In the following, we report experiments by training with the ground-truth captions in the datasets we use,
by finetuning our proposed model (Section~\ref{app:subsec:finetuning}),
and by demonstrating the advantages of MCQS on multi-captioned data (Section~\ref{app:subsec:app-mcqs-train}).

\subsection{Finetuning with ground-truth captions}
\label{app:subsec:finetuning}
We show that our proposed methodology can be used as a 
pretraining step.
Here, we experiment with initializing a model trained with automatic captions, and finetuning with ground-truth captions to further improve the performance.
Table~\ref{tab:finetuning} summarizes the results.
The bottom gray lines compare finetuning the model with ground-truth captions
(i) from CLIP initialization (rows with WiT+GT data), or (ii) from pretraining with our method (last row with WiT+GT+PL data).
This comparison highlights the benefits of using our proposed methodology as a pretraining
step, as it leads to further improvement in performance on the target datasets.
We note that when we train with the ground truth, we keep all hyperparameters the same for both (i) finetuning from CLIP initialization or (ii) finetuning from our pretraining with pseudolabels. 

\subsection{Multi-caption training on MSR-VTT}
\label{app:subsec:app-mcqs-train}
MSR-VTT videos come with 20 ground-truth captions per video.
Therefore, in the fully-supervised setting,
we can employ our MCQS approach for training.
In Table~\ref{tab:app-mcqs-train}, we show that using all ground-truth captions at a time
with MCQS improves over using a single caption randomly sampled at each training iteration.

\setlength{\tabcolsep}{5pt}
\begin{table}
\centering
\resizebox{0.99\linewidth}{!}{
\begin{tabular}{lll|cc|cc|cc}
    \toprule
    && Vision & \multicolumn{2}{c|}{ActivityNet} & \multicolumn{2}{c|}{MSR-VTT} & \multicolumn{2}{c}{MSVD} \\
    Method & Data & backbone & R@1 & R@5 & R@1 & R@5 & R@1 & R@5 \\
    \midrule
    \rowcolor{badcolor}
    CLIP & WiT & ViT-B/16 & 23.8 & 50.0 & 33.9 & 57.3 & 38.5 & 64.6 \\
    \midrule
    \rowcolor{goodcolor}
    Ours (Self) & WiT+PL & ViT-B/16 & 29.7 & 57.1 & 39.0 & 64.6 & 42.5 & 70.1 \\
    \rowcolor{goodcolor}
    Ours (Comb.) & WiT+PL & ViT-B/16  &  {30.6} & {57.9}  & {39.2} & {65.1} & {44.6} & {71.8} \\
    \midrule
    \gray{GT} &  \gray{WiT+GT} & \gray{ViT-B/16} & \gray{36.4} & \gray{66.5} & \gray{42.9} & \gray{70.9} & \gray{43.4} & \gray{74.3}  \\
    \gray{GT w/ QS} &  \gray{WiT+GT} & \gray{ViT-B/16} & \gray{35.1} & \gray{64.9} &  \gray{44.0} & \gray{70.5} & \gray{46.0} & \gray{73.9} \\
    \gray{GT w/ QS} &  \gray{WiT+GT+PL} & \gray{ViT-B/16} & \gray{\textbf{38.3}} & \gray{\textbf{68.8}} & \gray{\textbf{45.4}} & \gray{\textbf{72.4}} & \gray{\textbf{47.0}} & \gray{\textbf{75.0}} \\
    \bottomrule
  \end{tabular}
  }
    \caption{
    {\textbf{Fully-supervised setting:}~
    Comparison of Baseline, Ours, and training with Ground Truth (GT) captions. PL denotes training with the dataset videos without ground truth labels. The last row shows the results obtained by fine-tuning the \textit{Ours (Comb.)} model from Table~\ref{tab:combining-datasets}.
  }}
  \label{tab:finetuning}
\end{table}

\setlength{\tabcolsep}{3pt}
\begin{table}
\centering
\resizebox{0.80\linewidth}{!}{
\begin{tabular}{l|cc|cc}
    \toprule
    Caption & \multicolumn{2}{c|}{Temporal pooling}  &\multicolumn{2}{c}{MSR-VTT} \\
    pooling & train & eval & R@1 & R@5 \\
    \midrule
    CLIP~\cite{clip2021} & - & QS & 23.8 & 50.0 \\
    \midrule
    \gray{Random(GT)} & \gray{mean} & \gray{QS} & \gray{42.9} & \gray{70.9} \\
    \gray{Mean(GT)} & \gray{MCQS} & \gray{QS} & \gray{\textbf{44.9}} & \gray{\textbf{73.3}} \\
    \bottomrule
  \end{tabular}
  }
  \caption{\textbf{Multi-caption query-scoring training on MSR-VTT:}~Comparison of using a random single ground truth caption versus multiple ground truth captions at a time.
  }
  \label{tab:app-mcqs-train}
\end{table}

\section{Alternative methods}
\label{app:sec:baselines}
\noindent\textbf{Retrieving nearest-neighbor caption.}~~One interesting question is whether we 
need a captioner model to obtain frame captions. Given that there exists
a joint space between images and text through CLIP,
an alternative approach would be to retrieve
the closest text embedding from a large image-text gallery
by querying with the video frame embedding
(similar in spirit to \cite{nagrani2022}).
We performed this baseline
experiment using
the Google Conceptual Captions~\cite{sharma-etal-2018-conceptual}
dataset as the image-text gallery source, which was also the ClipCap training set \cite{ClipCap}. In a similar fashion to our previous experiments, (i) we extract 10 frames, (ii) retrieve a caption for each frame, and (iii) compute CLIPScore and filter them accordingly. In Table~\ref{tab:app-nn-caption},
we show that the retrieved captions can also be used to outperform the zero-shot
baseline. However, as will be seen in the next section,
the retrieved captions
appear to be less similar to the ground truth text
than with ClipCap or BLIP (Table~\ref{tab:app-nn-caption}).
\setlength{\tabcolsep}{6pt}
\begin{table}
\centering
\resizebox{0.99\linewidth}{!}{
\begin{tabular}{l|ll|ll}
    \toprule
    & \multicolumn{2}{c|}{MSR-VTT} & \multicolumn{2}{c}{MSVD} \\
    & R@1 & R@5 & R@1 & R@5 \\
    \midrule
    CLIP baseline~\cite{clip2021} & 32.8 & 55.7 & 39.4 & 64.6 \\
    \midrule
    Ours w/ OFA~\cite{wang2022ofa} & 33.6      & 59.2     & \underline{41.1} & 67.4 \\
    Ours w/ ClipCap~\cite{ClipCap} & 34.7 & 59.8 & 40.6 & 68.9 \\
    Ours w/ BLIP~\cite{li2022blip} & \textbf{35.8} & 60.6 & \textbf{41.1} & \textbf{69.1} \\
    \midrule
    Ours w/ NN-CC & 35.4 & \textbf{61.1} & 40.2 & 66.9 \\
    \bottomrule
  \end{tabular}
  }
  \caption{\textbf{Retrieving nearest neighbor caption from an image-text dataset:} Instead of using a captioner, we experiment with retrieving the captions from the Conceptual Captions~\cite{sharma-etal-2018-conceptual} dataset, using the frame embedding as query (NN-CC) and obtain comparable performance to other captioners.
  }
  \label{tab:app-nn-caption}
\end{table}


\setlength{\tabcolsep}{6pt}
\begin{table}
\centering
\resizebox{0.99\linewidth}{!}{
\begin{tabular}{ll|ll|ll|ll}
    \toprule
    && \multicolumn{2}{c|}{ActivityNet} & \multicolumn{2}{c|}{MSR-VTT} & \multicolumn{2}{c}{MSVD} \\
    & Text enc. & R@1 & R@5 & R@1 & R@5 & R@1 & R@5 \\
    \midrule
    \multicolumn{2}{l|}{CLIP baseline~\cite{clip2021}} & \textbf{23.4} & \textbf{49.3} & \textbf{32.8} & \textbf{55.7} & \textbf{39.4} & \textbf{64.6} \\
    \midrule
    ClipCap~\cite{ClipCap}  & CLIP   & 7.7  & 20.6 & 13.2 & 28.5 & 18.3 & 34.8 \\
    ClipCap~\cite{ClipCap}  & S-BERT & 9.3  & 26.3 & 16.4 & 34.3 & 20.4 & 44.4 \\
    \midrule
    BLIP~\cite{ClipCap}     & CLIP   & 10.6 & 28.5 & 15.8 & 33.3 & 25.7 & 47.6 \\
    BLIP~\cite{li2022blip}  & S-BERT & 13.1 & 32.3 & 18.1 & 39.0 & 28.5 & 52.3 \\
    \bottomrule
  \end{tabular}
  }
  \caption{\textbf{Captioning bottleneck with text-to-text retrieval:} We experiment with retrieving videos by representing them with the text embedding of the extracted captions. This results in lower performance than the CLIP baseline. We present performances with two different text encoders, the CLIP~\cite{clip2021} text encoder and Sentence-BERT~\cite{reimers-2019-sentence-bert} (S-BERT). See text for more details.
  }
  \label{tab:app-caption-bottleneck}
\end{table}

\noindent\textbf{Captioning bottleneck with text-to-text retrieval.}~~Another baseline we design is to use the captions
directly at test time without fine-tuning CLIP.
This constitutes an information bottleneck
where the video is embedded only into a text,
as opposed to a high-dimensional embedding space.
To determine the nearest video given a text query,
we use the previously extracted captions with ClipCap and BLIP.
To represent a given video, (i)~we embed the 10 extracted captions with Sentence-BERT~\cite{reimers-2019-sentence-bert} (S-BERT),
(ii)~select the two with the highest CLIPScore~\cite{hessel2021clipscore}, and
(iii)~average their embeddings.
We then compare a text query (also embedded with S-BERT) with this video representation
using cosine similarity.
In Table~\ref{tab:app-caption-bottleneck}, we summarize
the results.
Of the two text encodings tested, S-BERT performs better than CLIP text encoder as S-BERT was intentionally trained to detect similar sentences.
However, even the best performing caption bottleneck
(i.e., BLIP with S-BERT) obtains worse results
than the zero-shot CLIP baseline.
The poor performance
of this caption-based retrieval approach
suggests that captions are not sufficient
to be used directly for retrieval, but they
can instead provide a supervision signal for training.

\noindent\textbf{Text summarization.}
As mentioned in
\if\sepappendix1{Section~4.5} \else{Section~\ref{subsec:limitations}} \fi
of the main paper, we explored using a text summarization model to combine multiple captions in a given video, and our attempts led to inconsistent results, as seen in Table~\ref{tab:app-summarization}. We experimented with summarizing the 10 captions from the two captioners, (Summ(10C) for ClipCap and Summ(10B) for BLIP) and summarizing the filtered and combined 4 captions (Summ(2C+2B)). To summarize the captions, we use the Ada language model hosted in OpenAI.
We empirically find that it helps to prepend a randomly sampled raw caption to the summary, potentially because we obtain a longer caption with both local and global information
(i.e., results in Table~\ref{tab:app-summarization} improve when the prepend column is not empty, e.g., 37.5 vs 35.9).

\setlength{\tabcolsep}{3pt}
\begin{table} 
\centering
\resizebox{0.99\linewidth}{!}{
\begin{tabular}{ll|cc|cc}
    \toprule
    & & \multicolumn{2}{c|}{MSR-VTT} & \multicolumn{2}{c}{MSVD} \\
    Prepend & Summary & R@1 & R@5 & R@1 & R@5 \\
    \midrule
    2C + 2B       &  -          & 36.5 & 61.5 & \textbf{41.7} & \textbf{70.0} \\
    \midrule
    -       & Summ(10C)         & 32.1 & 58.0 & 39.4 & 64.7 \\
    10C     & Summ(10C)         & 33.6 & 58.8 & 40.3 & 65.8 \\
    -       & Summ(10B)         & 33.7 & 59.2 & 40.6 & 68.0 \\
    10B     & Summ(10B)         & 34.4 & 59.1 & 41.0 & 69.0 \\
    \midrule
    -       & Summ(2C + 2B)     & 35.9 & 60.9 & 40.8 & 68.8 \\
    2C + 2B & Summ(2C + 2B)     & \textbf{37.5} & \textbf{62.2} & 38.6 & 69.4 \\
    \bottomrule
  \end{tabular}
  }
  \caption{\textbf{Text summarization results:}
  Results when summarizing the 10 available captions or the Top 2 from each captioner (2C+2B). We explore two variants, training with only the summary (prepend empty), or training with the concatenation of a random caption and the summary. We do not obtain consistent improvements.
  }
  \label{tab:app-summarization}
\end{table}

\section{Additional evaluations}
\label{app:sec:evaluations}
In this section, we report
cross-dataset evaluations (Section~\ref{app:subsec:crossdataset}),
multi-caption evaluation on ActivityNet (Section~\ref{app:subsec:app-mcqs-eval}), and
performance metrics for video-to-text retrieval (Section~\ref{app:subsec:v2t}).

\subsection{Cross-dataset evaluation}
\label{app:subsec:crossdataset}
As mentioned in
\if\sepappendix1{Section~4.2} \else{Section~\ref{subsec:ablation}} \fi
of the main paper,
we report cross-dataset evaluations.
In Table~\ref{tab:app-cross-dataset-eval},
we use the models
trained with multi-caption query scoring, where the diagonal corresponds to the second-last row
of
\if\sepappendix1{Table~5} \else{Section~\ref{tab:combining-datasets}} \fi
(training and evaluating on the same dataset). Interestingly, the performance of MSR-VTT training and evaluating 
on ActivityNet is almost as good as training with ActivityNet videos. Furthermore, 
models trained only on MSVD perform poorly on all datasets
(including itself),
given its small size.
\setlength{\tabcolsep}{3pt}
\begin{table}
\centering
\resizebox{0.99\linewidth}{!}{
\begin{tabular}{l|cc|cc|cc}
    \toprule
    \multirow{2}{*}{\backslashbox{Train}{Eval}}  & \multicolumn{2}{c|}{ActivityNet} & \multicolumn{2}{c|}{MSR-VTT} & \multicolumn{2}{c}{MSVD} \\
    & R@1 & R@5 & R@1 & R@5 & R@1 & R@5 \\
    \midrule
    \rowcolor{badcolor}
    CLIP~\cite{clip2021} & 23.8 & 50.0 & 33.9 & 57.3 & 38.5 & 64.6 \\
    \midrule
    
    ActivityNet     & \cellcolor{regularcolor}29.7 & \cellcolor{regularcolor}57.0 & 38.4 & 62.7 & 43.3 & 69.7 \\
    MSR-VTT         & 29.5 & 56.7 & \cellcolor{regularcolor}39.0 & \cellcolor{regularcolor}64.6 & 43.5 & 69.2 \\
    MSVD            & 28.8 & 55.4 & 37.9 & 62.7 & \cellcolor{regularcolor}42.5 & \cellcolor{regularcolor}70.0 \\
    \midrule
    \rowcolor{goodcolor}
    Combined & \textbf{30.6} & \textbf{57.9}  & \textbf{39.2} & \textbf{65.1} & \textbf{44.5} & \textbf{71.8} \\
    \bottomrule
  \end{tabular}
  }
  \caption{\textbf{Cross-dataset evaluation:}
  Diagonal is training and evaluating on the same dataset (Table 5, Self row, in the main paper). Training with MSVD leads to lowest performance (smallest dataset among three).
  Note that we train three MSVD models with different seeds and report the mean of the recalls.}
  \label{tab:app-cross-dataset-eval}
\end{table}

\subsection{Multi-caption evaluation on ActivityNet}
\label{app:subsec:app-mcqs-eval}
To evaluate ActivityNet in all the experiments in the paper, we concatenate all the ground-truth captions available for a video and generate a text query as in~\cite{clip4clip2021, Zhang2018Cross-Modal,gabeur2020mmt}. Instead of concatenating the multiple captions to form a single text query,
we can use all the available descriptions of a video as text queries and evaluate using our multi-caption query scoring method.
In Table~\ref{tab:app-mcqs-eval},
we observe further improvements with this
approach.

\setlength{\tabcolsep}{3pt}
\begin{table}
\centering
\resizebox{0.70\linewidth}{!}{
\begin{tabular}{ll|cc}
    \toprule
    & & \multicolumn{2}{c}{ActivityNet} \\
    Method & Eval & R@1 & R@5 \\
    \midrule
    CLIP~\cite{clip2021} & QS & 23.8 & 50.0 \\
    \midrule
    Ours (Combined) & QS & 30.6 & 57.9 \\
    Ours (Combined) & MCQS & \textbf{31.7} & \textbf{58.8} \\
    \bottomrule
  \end{tabular}
  }
  \caption{\textbf{Multi-caption query-scoring evaluation on ActivityNet:}~We compare evaluating with query-scoring (QS) with a single text query per video (concatenating descriptions), with multiple-caption query-scoring.
  }
  \label{tab:app-mcqs-eval}
\end{table}

\subsection{Video-to-text retrieval metrics}
\label{app:subsec:v2t}
In the main paper, we only report text-to-video retrieval metrics. Here, in Table~\ref{tab:app-v2t}, we report the
\textit{video}-to-text metrics. We see that our method also improves over the baseline on these metrics.

\setlength{\tabcolsep}{3pt}
\begin{table}
\centering
\resizebox{0.99\linewidth}{!}{
\begin{tabular}{l|cc|cc|cc}
    \toprule
    &  \multicolumn{2}{c|}{ActivityNet} & \multicolumn{2}{c|}{MSR-VTT} & \multicolumn{2}{c}{MSVD} \\
    Method & R@1 & R@5 & R@1 & R@5 & R@1 & R@5 \\
    \midrule
    \rowcolor{badcolor}
    CLIP baseline~\cite{clip2021} & 21.5 & 45.6 & 32.3 & 56.3 & 35.4 & 62.4 \\
    \midrule
    \rowcolor{goodcolor}
    Ours (Self)     & 28.5     & \textbf{56.0}   & \textbf{36.5} & 64.0 & 40.0 & 69.7 \\
    \rowcolor{goodcolor}
    Ours (Combined) & \textbf{28.7} & 55.9  & 36.4 & \textbf{66.4} & \textbf{41.6} & \textbf{70.5} \\
    \bottomrule
  \end{tabular}
  }
  \caption{\textbf{Video-to-text retrieval metrics:} Our method (2C+2B trained with MCQS and evaluated with QS) also improves the CLIP baseline on video-to-text retrieval metrics.}
  \label{tab:app-v2t}
\end{table}

\section{Analysis on selecting captions and combining multiple captioners}
\label{app:sec:combining-captioners}
As mentioned in
\if\sepappendix1{Section~4.2} \else{Section~\ref{subsec:ablation}} \fi
of the main paper,
we provide further analysis 
about the source of captions from multiple frames
and multiple captioners.

\noindent\textbf{Quantitative results.}
One way to check the assumption that selecting the best captions is removing noisy captions is to compare the captions with the ground truth. In Table~\ref{tab:app-caption-metrics}, 
we compare the extracted captions with the ground truth with two metrics:
METEOR and CLIPScore.
However, unlike in the main paper, here we compute the CLIPScore
between the two texts (extracted and ground-truth captions),
rather than between visual and text embeddings.
The results motivate the top-2 selection instead of using all 10 captions.
We also show the maximum (Max), which corresponds to the comparison of the ground truth with each of the 10 captions individually and selecting the one with the highest score, as a way to give an upperbound on this score assuming a perfect selection method (note that this requires access to the ground truth).
%
We observe that retrieving nearest neighbor captions has the least similarity with the ground-truth text. 

\setlength{\tabcolsep}{6pt}
\begin{table}
\centering
\resizebox{1.0\linewidth}{!}{
\begin{tabular}{ll|cc|cc|cc}
    \toprule
    &  &  \multicolumn{2}{c|}{ActivityNet} & \multicolumn{2}{c|}{MSR-VTT} & \multicolumn{2}{c}{MSVD} \\
    & \# capt. & M & T-CS & M & T-CS & M & T-CS \\
    \midrule
    \multirow{3}{*}{NN} & 10    & - & - & \white{0}6.9 & 77.9 & \white{0}7.2 & 70.1 \\
                        & Top 2 & - & - & \white{0}7.8 & 79.6 & \white{0}8.5 & 72.1 \\
                        & \gray{Max} & \gray{-} & \gray{-} & \gray{12.0} & \gray{85.9}  & \gray{14.9} & \gray{78.4} \\
    \midrule
    \multirow{3}{*}{C} & 10      & 15.3  & 70.3 & \white{0}9.1      & 81.9   & \white{0}9.3  & 73.0\\
    & Top 2  & 16.6  & 71.2 & 10.4     & 82.2   & 10.7 & 74.2 \\
    & \gray{Max}& \gray{26.1} & \gray{78.4} & \gray{14.9} & \gray{88.6} & \gray{17.9} & \gray{80.6}  \\
    \midrule
    \multirow{3}{*}{B} & 10      & 17.2  & 74.0 & 20.2     & 85.6  & 21.0    & 79.0 \\
    & Top 2  & 17.7  & 74.4 & 20.5     & 86.4  & 21.7    & 79.8 \\
    & \gray{Max} & \gray{28.0} & \gray{80.4}& \gray{27.5} & \gray{91.9}  & \gray{31.8} & \gray{84.6} \\
    \bottomrule
  \end{tabular}
  }
  \caption{\textbf{Comparing automatic captions to ground-truth text:}
  We compare the extracted captions from Nearest Neighbour (NN), ClipCap (C) and BLIP (B) approaches to ground-truth video captions, with METEOR (M)~\cite{banerjee-2005-meteor} and Text CLIPScore (T-CS)~\cite{hessel2021clipscore} metrics.
  When we evaluate 10 or 2 captions, we compute the metrics individually for each caption and report the average. For the maximum, we compute the metrics for all the 10 captions and select the one with the highest score.
  Retrieving nearest neighbour captions have the least similarity with the ground truth text.
  Filtering captions with CLIPScore (Top 2) improves all metrics.
  }
  \label{tab:app-caption-metrics}
\end{table}

\noindent\textbf{Different CLIPScore distributions.}
As seen in Figure~\ref{fig:app-cs-kde}, ClipCap and BLIP captions have different CLIPScore distributions, with $\mu$ being higher for ClipCap, perhaps due to the CLIP backbone. If we were to select the best 4 captions out of the 20 available ones, we would be selecting ClipCap captions more often than BLIP captions.

\begin{figure}
    \includegraphics[width=.99\linewidth]{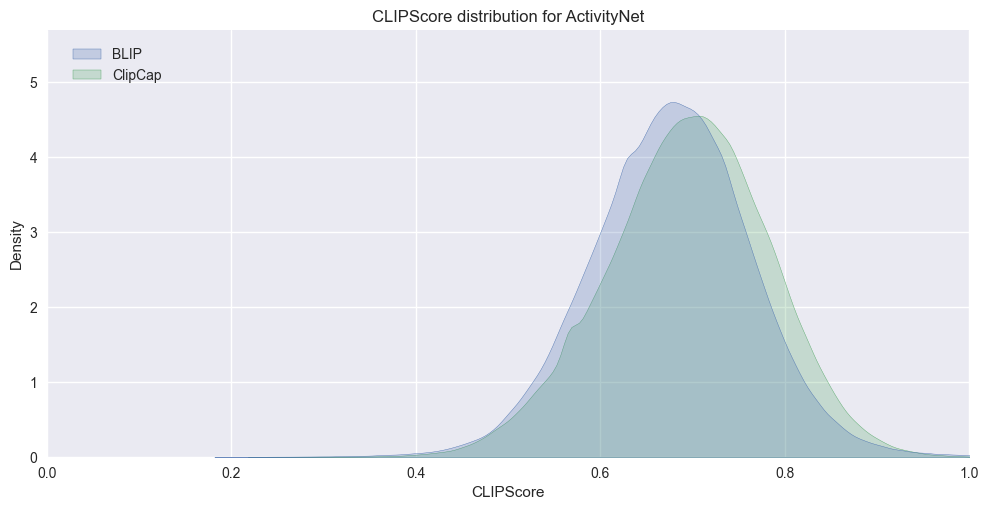}
    \includegraphics[width=.99\linewidth]{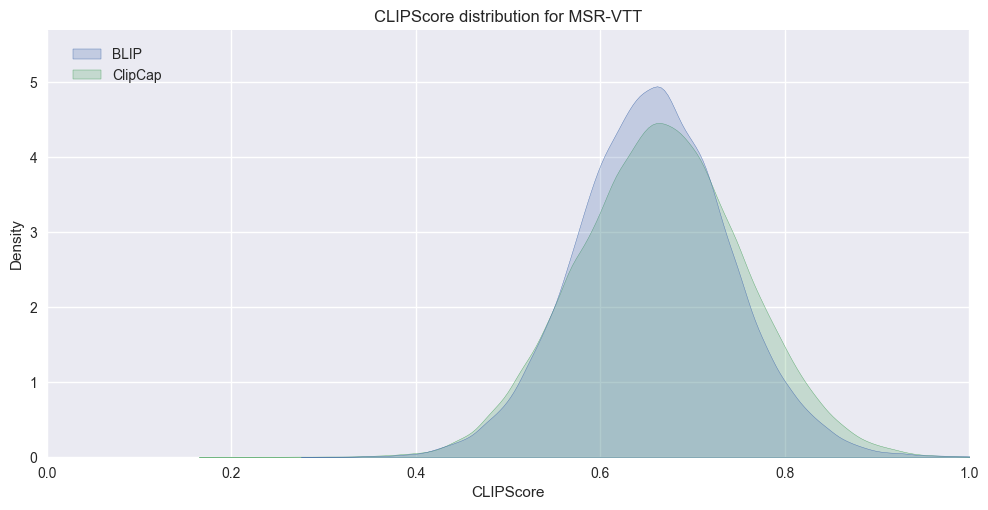}
    \includegraphics[width=.99\linewidth]{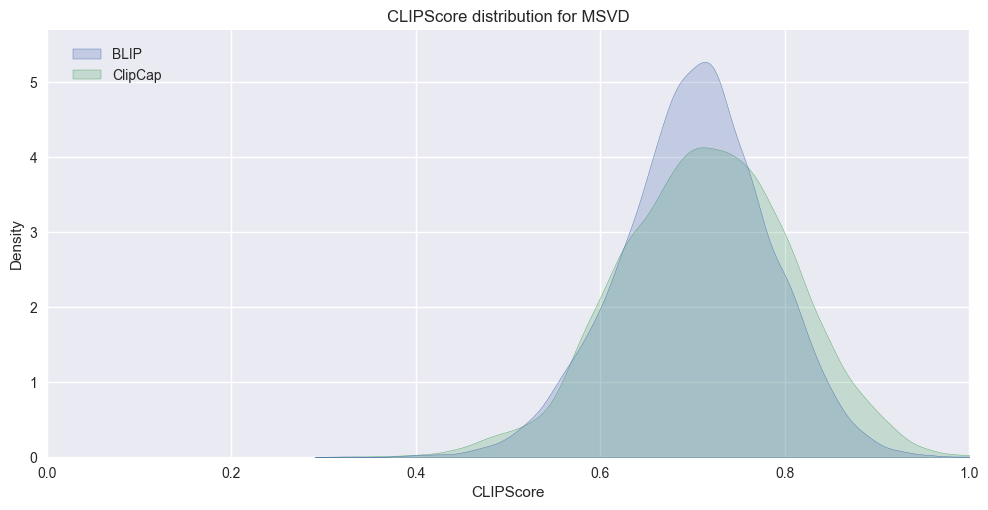}
    \caption{\textbf{CLIPScore kernel density estimate:} We plot the CLIPScore distribution for three datasets, and both models (ClipCap and BLIP). CLIPScore is higher for ClipCap than for BLIP, potentially because of the CLIP backbone.}
    \label{fig:app-cs-kde}
\end{figure}

\noindent\textbf{Top 4 of all the captions.}~~We see in Figure~\ref{fig:app-combining-captioners} that combining captions from different captioners is better than using only ClipCap or BLIP. Out of the two alternatives: (i) selecting top 4 of the 20 combined set of captions, (ii) selecting top 2 from each captioner, option (ii) leads to better results.

\begin{figure}
    \includegraphics[width=.99\linewidth]{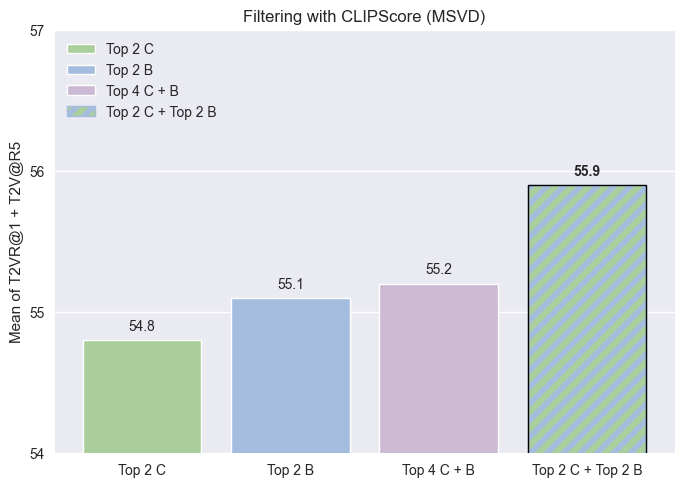}
    \caption{\textbf{Combining captioners:} We compare 4 different strategies: selecting 2 from 10 ClipCap captions, selecting 2 from 10 BLIP captions, selecting Top 4 from the 20 combined captions, selecting Top 2 from each captioner. We highlight the best performance with a black border.
    }
    \label{fig:app-combining-captioners}
\end{figure}

\noindent\textbf{Number of different frames.}~~When we select Top 2 from one captioner, our captions come from only two frames. In Table~\ref{tab:app-frames}, we see statistics of the amount of different frames when combining Top 2 of ClipCap with Top 2 of BLIP. It can be seen that
only around 7\% of the time the top captions from both captioners come from the exact two frames. More than 44\% of the time there is a frame in common with the two captioners. 
Finally, most frequently, 4 \textit{different} frames are selected from the 10 possible frames: 2 from each captioner.

\setlength{\tabcolsep}{6pt}
\begin{table}
\centering
\resizebox{0.89\linewidth}{!}{
\begin{tabular}{l|c|c|c}
    \toprule
    Dataset & 4 frames & 3 frames & 2 frames \\
    \midrule
    ActivityNet & 47.4\% & 45.4\% & 7.2\%   \\
    MSR-VTT     & 48.5\% & 44.2\% & 7.3\%   \\
    MSVD        & 47.4\% & 44.8\% & 7.8\%   \\

    \bottomrule
  \end{tabular}
  }
  \caption{\textbf{Different frames:} When using C+B Top-2 (4 captions), about 
  47\% of the videos have captions from 4 different frames, and around 45\% of the videos have captions from 3 different frames (i.e., the two captioners pick the same one frame in their top rankings). Finally, there are roughly 7\% of the videos where both captioners select the same two frames. In these cases, multiple captions can still be useful to provide data augmentation.
  }
  \label{tab:app-frames}
\end{table}

\noindent\textbf{Repetitive captions.}~~One other benefit of filtering the captions is that we are left with a set of less repetitive captions. See
Figure~\ref{fig:app-unique-captions} for the percentage of unique captions when using 10 captions and Top 2 captions.
We also check that there are less than 1\% of overlapping captions between the two captioners in any of the three datasets. This is yet another reason that motivates us to use different captioners and obtain more diverse and rich captions.

\begin{figure}
    \centering
    \includegraphics[width=.99\linewidth]{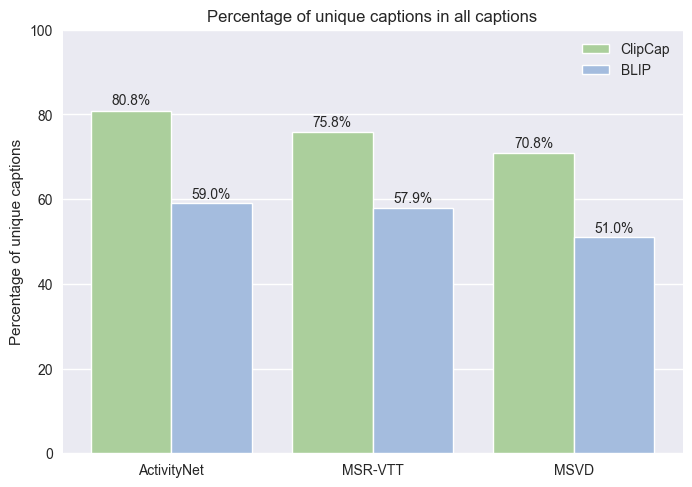}
    \includegraphics[width=.99\linewidth]{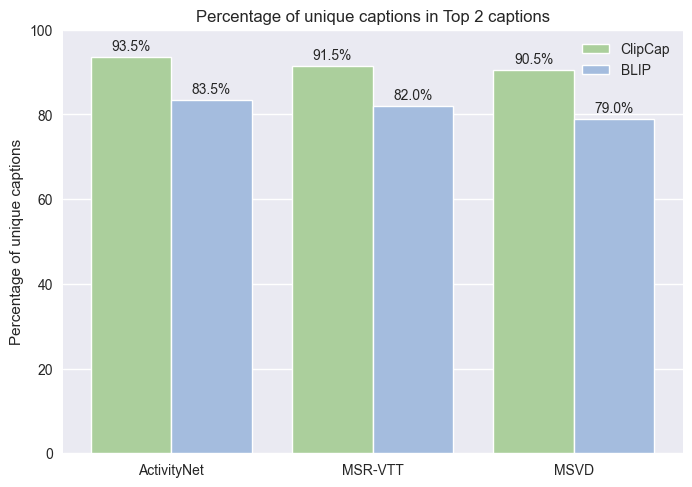}
    \caption{\textbf{Percentage of unique captions:}
    We make statistics about the percentage of unique extracted captions 
    within a video (top: for all 10 captions, bottom: for the best 2 captions).
    We observe that
    BLIP captions are more diverse, and ClipCap ones are a bit more repetitive.
    }
    \label{fig:app-unique-captions}
\end{figure}

\setlength{\tabcolsep}{6pt}
\begin{table}
\centering
\resizebox{1.0\linewidth}{!}{
\begin{tabular}{ll|cc|ll}
    \toprule
    & Caption & \multicolumn{2}{c|}{Temporal pooling} & \multicolumn{2}{c}{MSVD} \\
    Captioners & pooling & train & eval & R@1 & R@5 \\
    \midrule
    \multicolumn{2}{l|}{\multirow{2}{*}{CLIP baseline~\cite{clip2021}}} & - & mean & 39.4 & 64.5 \\
    \multicolumn{2}{l|}{\multirow{2}{*}{}} & - & QS & 38.5 & 64.5 \\
    \midrule
    C + B & Rand(4) & mean  & mean  & 41.7         & \textbf{70.0}  \\
    C + B & Mean(4) & MCQS  & QS    & 42.5         & \textbf{70.0} \\
    \midrule
    C + B + O & Rand(6) & mean  & mean  & 41.8         & 69.2  \\
    C + B + O & Mean(6) & MCQS  & QS    & \textbf{42.8}& 68.5  \\
    \bottomrule
  \end{tabular}
  }
  \caption{\textbf{Combining with OFA:}
  We experiment with combining three captioners, i.e.,
  using a total of 6 captions by selecting top 2 from
  each of the ClipCap (C), BLIP (B), and OFA (O) captioners.
  While the R@1 metric improves when adding a third captioner, we see no further improvement in R@5.
  }
  \label{tab:app-combining-ofa}
\end{table}

\noindent\textbf{Beyond two captioners.}~~We explore using three different captions by combining ClipCap (C), BLIP (B) and OFA (O) in Table~\ref{tab:app-combining-ofa}. The results do not bring consistent improvements
in both metrics (better R@1, worse R@5), possibly because
OFA performance alone is not as effective compared
to BLIP.

\section{\new{Implementation details for the BLIP initialization experiment}}
\label{app:sec:blip-implementation-details}
\new{
We here explain the BLIP implementation details of the backbone experiments in
\if\sepappendix1{Table~6.} \else{Section~\ref{tab:backbones}.} \fi
We train using a method akin to that of BLIP, where the Image-Text Contrastive (ITC) loss is denoted as our $L$ in Eq.~(\ref{eq:loss}).
For the Image-Text Matching (ITM) loss, we extend the encoder hidden states by the number of frames.
We train with 4 frames and evaluate with 8 frames.
We adopt the ViT-B/16 backbone for the image encoder
and the BERT architecture~\cite{devlin-etal-2019-bert} for the text encoder as in BLIP.
We train the model with a single NVIDIA RTX A600
using 4 frames, 
while evaluations are conducted using 8 frames
as in the original paper. 
}

\section{Additional qualitative results}
\label{app:sec:qualitative}

\noindent\textbf{Captioning.}
Similar to 
\if\sepappendix1{Figure~2} \else{Figure~\ref{fig:method}} \fi
of the main paper,
in Figure~\ref{fig:app-qualitative-captioners},
we provide more examples of captioning results
from both ClipCap and BLIP, together with their corresponding
CLIPScores when compared to the image embeddings.
In the third picture of the second video or in the first picture of the third video, we see that CLIPScore is low when the captions does not match the frame. In the last video, we see examples of a short video where all the frames look alike, and the extracted captions are the same or almost the same.

\noindent\textbf{Retrieval.}
To complement 
\if\sepappendix1{Figure~3} \else{Figure~\ref{fig:qualitative-results}} \fi
of the main paper, we provide additional qualitative results 
in Figure~\ref{fig:app-qualitative-t2v-results}
for the three datasets: ActivityNet (first two rows), MSR-VTT (middle two rows) and MSVD (last two rows).

\section{Data availability statement}
\label{app:data-availability}

We conducted experiments using three popular text-to-video retrieval public datasets, namely ActivityNet~\cite{krishna2017dense}, MSR-VTT~\cite{Xu2016msrvtt}, and MSVD~\cite{chen-2011-msvd}. The URLs to download the datasets are:

\begin{itemize}
\item \href{https://cs.stanford.edu/people/ranjaykrishna/densevid/}{ActivityNet}
\item \href{https://github.com/albanie/collaborative-experts/blob/master/misc/datasets/msvd/README.md}{MSR-VTT}
\item \href{https://www.mediafire.com/folder/h14iarbs62e7p/shared}{MSVD}
\end{itemize}
We complement them with our automatic caption labels and will release these along with our code and pretrained models.

\begin{figure*}
    \centering
    \includegraphics[width=.99\textwidth]{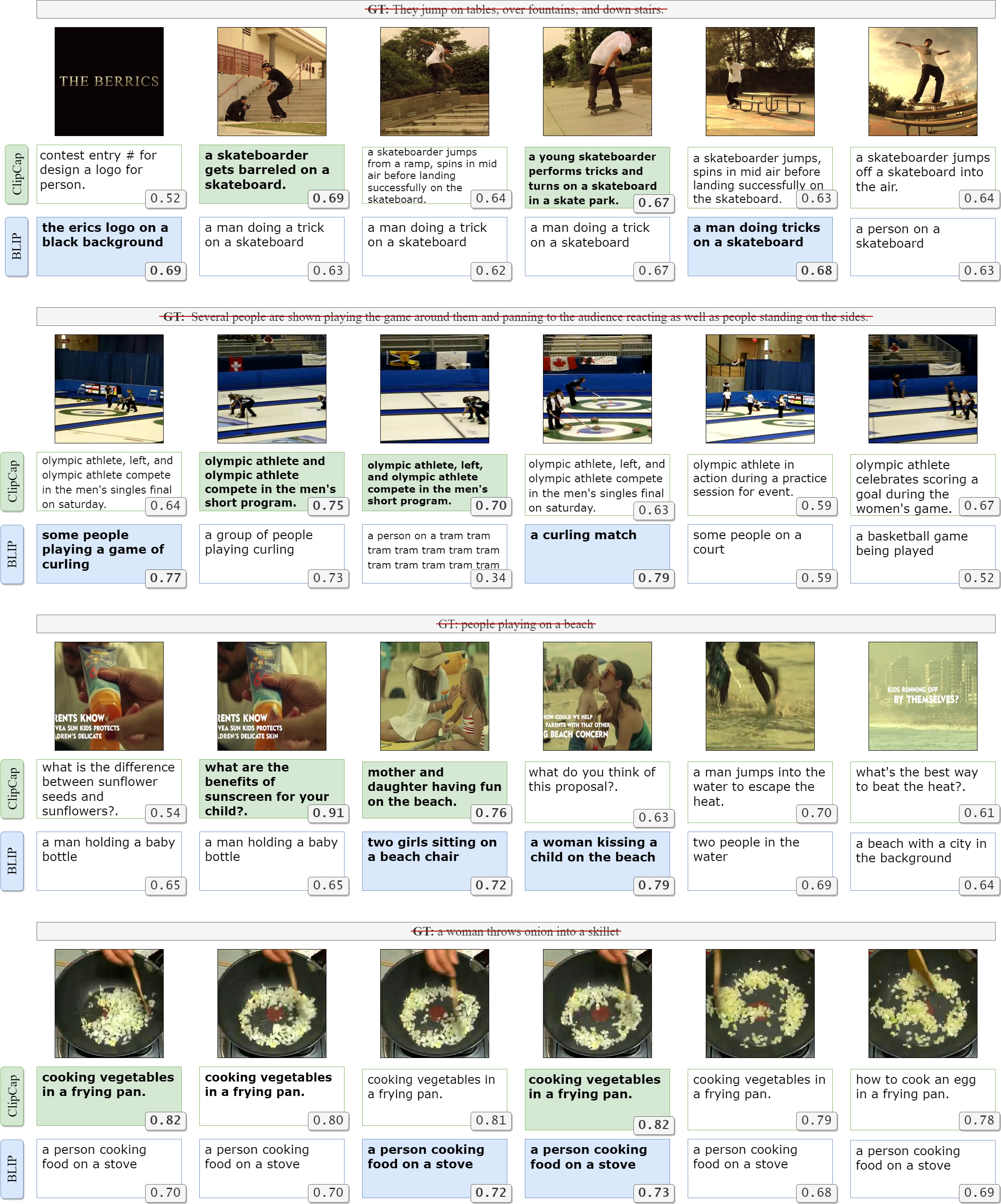}
    \caption{\textbf{Qualitative results for captioning:} We visualize further captioning results by both \colorbox{clipcapcolor}{ClipCap} and \colorbox{blipcolor}{BLIP} as in
    \if\sepappendix1{Figure~2}
    \else{Figure~\ref{fig:method}} \fi
    of the main paper.
    First two rows are on ActivityNet, third row on MSR-VTT, last row on MSVD.
    }
    \label{fig:app-qualitative-captioners}
\end{figure*}

\begin{figure*}
    \centering
    \includegraphics[width=.90\textwidth]{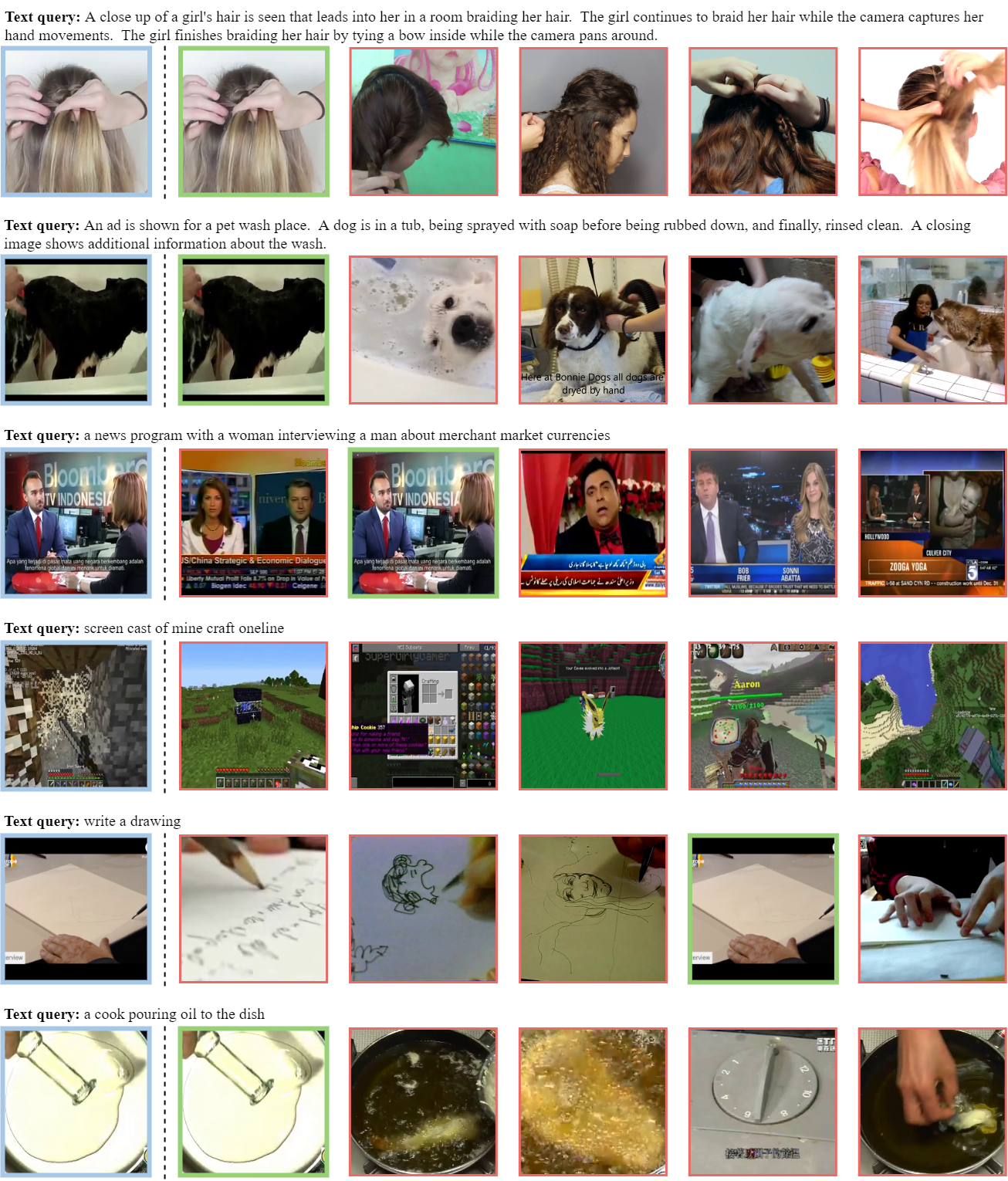}
    \caption{\textbf{Qualitative text-to-video retrieval results:}
    Above, video retrieval results for our best model (Combined) are shown. The examples belong to the test sets of ActivityNet (first two rows), MSR-VTT (third and fourth rows), and MSVD
    (last two rows). Each example is shown with the text query, the ground-truth video (first column, blue border), and the top 5 retrieved videos from the gallery. Every video is only displayed using the middle frame, with a green border if it matches the ground-truth video, or a red border otherwise. Overall, all the retrieved videos have similar semantic meaning with the text query, even in cases where the correct video is not retrieved at the first rank.
    }
    \label{fig:app-qualitative-t2v-results}
\end{figure*}

\end{document}